\documentclass[manuscript]{acmart}

\AtBeginDocument{%
  \providecommand\BibTeX{{%
    \normalfont B\kern-0.5em{\scshape i\kern-0.25em b}\kern-0.8em\TeX}}}

\copyrightyear{2021} 
\acmYear{2021} 
\setcopyright{acmlicensed}\acmConference[MEMSYS 2021]{The International Symposium on Memory Systems}{September 27--30, 2021}{Washington DC, DC, USA}
\acmBooktitle{The International Symposium on Memory Systems (MEMSYS 2021), September 27--30, 2021, Washington DC, DC, USA}
\acmPrice{15.00}
\acmDOI{10.1145/3488423.3519324}
\acmISBN{978-1-4503-8570-1/21/09}

\usepackage{balance}
\usepackage{subcaption}

\begin{document}

\title{Learning to Rank Graph-based Application Objects on Heterogeneous Memories}

\author{Diego Moura}
\affiliation{%
  \institution{Federal University of Bahia}
  \city{Salvador}
  \country{Brazil}
}
\email{diego.braga@ufba.br}

\author{Daniel Mossé}
\affiliation{%
  \institution{University of Pittsburgh}
  \city{Pittsburgh}
  \country{USA}}
\email{mosse@cs.pitt.edu}

\author{Vinicius Petrucci}
\affiliation{%
 \institution{Federal University of Bahia \& University of Pittsburgh}
 \city{Pittsburgh}
 \country{USA}}
\email{vpetrucci@pitt.edu}

\renewcommand{\shortauthors}{Diego Moura, Daniel Mossé and Vinicius Petrucci}

\begin{abstract}
  Persistent Memory (PMEM), also known as Non-Volatile Memory (NVM), can deliver higher density and lower cost per bit when compared with DRAM. Its main drawback is that it is typically slower than DRAM. On the other hand, DRAM has scalability problems due to its cost and energy consumption. Soon, PMEM will likely coexist with DRAM in computer systems but the biggest challenge is to know which data to allocate on each type of memory. This paper describes a methodology for identifying and characterizing application objects that have the most influence on the application's performance using Intel Optane DC Persistent Memory. In the first part of our work, we built a tool that automates the profiling and analysis of application objects. In the second part, we build a machine learning model to predict the most critical object within large-scale graph-based applications. Our results show that using isolated features does not bring the same benefit compared to using a carefully chosen set of features. By performing data placement using our predictive model, we can reduce the execution time degradation by 12\% (average) and 30\% (max) when compared to the baseline's approach based on LLC misses indicator.
\end{abstract}




\maketitle

\section{Introduction}

Systems with heterogeneous memories are already a reality \citep{IntelOptaneTecnology,argone,joseph2019}. One of the main reasons for this possible trend is due to the scalability problems of DRAM \cite{Malladi2012,Shalf2010,Lee2009} and the strong demand for memory capacity arising from data intensive applications \cite{Redis,Memcached}.
If, on one hand, a heterogeneous system offers a lower price and has the potential for good performance, on the other hand, it adds complexity to memory management. For example, which objects to place in which of available memory tiers? To answer this and other questions, it is essential to characterize the behavior of application allocations to identify the most critical object (the hottest).

The characterization of an application can usually be done in two ways: \textit{online} and \textit{offline}. The main advantage of the \textit{online} approach is that no profiling is necessary and the application data is collected in real-time. 
However, it has the disadvantage of being a reactive and local approach that may not present a global view of the application. The \textit{offline} approach requires the collection of the application profile in advance for decision making. By requiring such a prior knowledge, it may provide a view of the entire application and one could potentially make better decisions.  Our work focuses on the second approach.

When profiling an offline application, two tasks are fundamental: (1) characterize individual objects and  their properties (also known as \textit{features}) and (2) perform the data placement. There are available tools \cite{Liu2013, VTune} that perform task 1 without changing the application, but not task 2. There are prior works that perform tasks 1 and 2, but require changing the application \cite{Wen2018}. Our tool is able both to profile the objects of an application and to perform data placement without the intervention of the developer. Concerning the features for each object, the most used methodology counts the number of cache misses in the last level cache (LLC) for each object and the object with the highest value usually has the highest priority to be allocated in the fastest memory (DRAM)  \cite{Wen2018,chen2020,Servat2017}. However, in a context where the Intel Optane and other non-volatile memory technologies are used, LLC misses alone is not sufficient to quantify the importance of an object, as we show in Section~\ref{sec:motivation}. 

Our work proposes \textit{Rank Hottest Object} (RHO), a machine learning model responsible for finding the hottest object considering a set of attributes. Given an application (source code or binary) our tool identifies the objects without changing the application, calculates metrics for each of them, and using our model ranks the object with the greatest potential for gain in execution time when allocated in the DRAM. We carefully analyzed hardware performance counters including the intensity of operations of loads with their respective latencies in different hierarchy levels of memory (L1, L2, L3, and DRAM), the intensity of operations of stores in L1, the ratio of TLB miss/hit, and the memory footprint (see Section~\ref{subsec:Predictive_modeling}).

We compare the performance of RHO with the methodology that is based on the number of LLC misses (the baseline). The results of the evaluation show that our predictive model  achieves application performance gains of up to 30\% and an average of 12\% when replacing the most sensitive object in the DRAM.

This paper makes the following contributions:

\begin{itemize}
    \item We present a characterization of real-world irregular memory access from graph-based applications using Intel Optane. Our characterization results show that Last Level Cache misses (LLCM) does not always deliver the best performance when using Intel Optane technology.
    \item We propose \textit{Rank Hottest Object} (RHO), a methodology based on Machine Learning for identifying the hottest object in an application exploring several metrics in a combined fashion.
    \item We quantify the effectiveness of RHO using a full real heterogeneous memory system and real applications/dataset. Our quantitative evaluation demonstrates the RHO hits the \emph{Top 1} object in 92\% of cases and improves the LLCM in execution time loss by 12\% on average. 
\end{itemize}

\section{Background}

\subsection{Graph Applications}
\label{section:graphs_application}

Because of the importance of analyzing graph applications
(e.g., social networks, maps, etc), GAPBS \cite{gapbs} applications have become very popular. These applications work by analyzing big data in a graph format, using data points as nodes and relationships as edges.  The following is a summary of the algorithms available on GAPBS:

\begin{itemize}
    \item \textbf{Betweenness Centrality (BC)}: measures the importance of a node in a graph. In social networks analysis, it is actively used for computing the user “influence” index. The vertex index reflects the fraction of shortest paths between all vertices that pass through a given vertex.
    \item \textbf{Breadth First Search (BFS)}: traverses a graph from the root node and explores all the neighbouring nodes. The BFS can be used in different scenarios such as locate all the nearest or adjacent nodes in a peer-to-peer network.
    \item \textbf{Connected Components (CC)}: identifies the maximal sets of vertices reachable from each other in an undirected graph.
    \item \textbf{PageRank (PR)}: ranks web pages based on the topology of a web graph. This algorithm was popularized through its use by Google Web Search.
    \item \textbf{Single Source Shortest Path (SSSP)}: calculates the shortest (weighted) path from a node to all other nodes in the graph.
    \item \textbf{Triangle Counting (TC)}: computes the total number of triangles in a graph. Triangles can be used for various tasks in real‐life networks such as spam filtering and community discovery.
\end{itemize}

In addition to working with large datasets, this type of application is characterized by having poor temporal and spatial locality which makes memory a performance bottleneck \cite{Ozdal2016}.
In our setup, memory demands for those applications range from 400 MB and 35 GB (more details in Section \ref{section:experimental_setup}). 

\subsection{Optane Memory}

Intel Optane DC Persistent Memory Module (or just ``Optane") is the current technology that comes closest to DRAM concerning performance. Furthermore, because PMEMs are packaged in DIMMS and are on the same bus/channel of the DRAM, it is not possible to achieve the theoretical maximum bandwidth for DRAM and PMEM when used together. Through this common bus, the CPU connects each of its two integrated memory controllers (iMC) to three channels. Communication between the iMC and the Optane is done with a granularity of 64 bytes. The communication between the Optane's internal controller and the banks of Optane Memory is done with a 256-byte granularity. This difference has consequences that will be 
discussed below. The Optane is available in three different capacities: 128 GB, 256 GB, and 512 GB. Since each CPU can host six Optane DIMMS, we can have a maximum of 3TB of PMEM. Given that most RAM available on the market does not exceed the capacity of 128GB DIMM, the Intel Optane already starts offering 4x more capacity.

In addition to the capacity of the Optane, three other features should be highlighted: latency, bandwidth, and asymmetry between read and write operations. The read latency of the Optane compared to DRAM is about 3x for random accesses and 2x for sequential access \cite{joseph2019}. The reason is that  adjacent requests are combined in a buffer before a store happens using the 256-byte internal granularity. This is a signal that if we coordinate the access to the PMEM, we can explore  locality, which can reduce latency and write-amplification. The store latency is difficult to measure because there is no mechanism to record when a store physically reaches the Optane (due to write caches and delayed writes). 

When comparing Optane's bandwidth with DRAM for a different number of threads (between 1 and 21 threads), prior work \cite{joseph2019} has shown that as the number of threads increases, the difference between the two technologies increases. For reading, DRAM achieves values above 100 GB/s while Optane achieves about 40 GB/s. For writing, the difference is even greater, about 80 GB/s for DRAM and around 14 GB/s for Optane when using 21 threads. Other researchers have highlighted \cite{Jian2020} that there are nuances when using this new technology, most notably (a) stores cost more (has higher latency) than loads, (b) access with request sizes less than 256 bytes tends to waste bandwidth, and (c) writing result in write amplification. In case (a), it is important to avoid allocations in the PMEM that have a lot of stores to minimize the impact on performance. In case (b), any load or store smaller than 256 bytes will have the same latency and the same bandwidth consumption as 256-byte access. So, if we can optimize these accesses, we can reduce this bandwidth waste. 

\begin{figure}[t]
    \centering
    \includegraphics[scale=0.33]{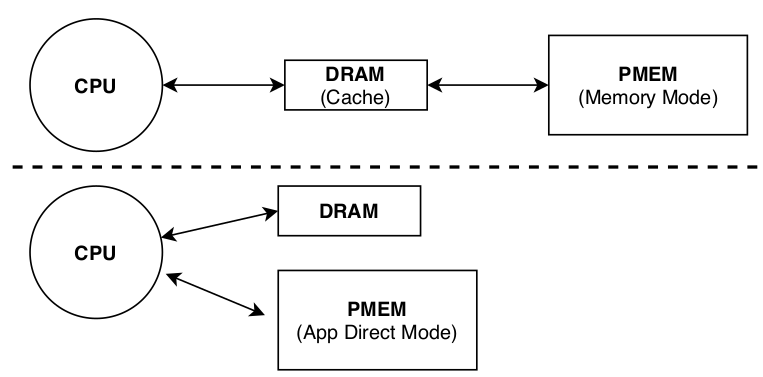}
    \caption{Memory Mode versus App Direct Mode.}
    \label{fig:memory_mode_vs_app_direct}
    \Description{A figure showing in the first part that in memory mode there is a heterogeneous architecture in the vertical format where the order of the organization is CPU, DRAM and then PMEM. In the second part there is a horizontal architecture, in which the CPU can directly access both DRAM as well as the PMEM.}
\end{figure}

\subsection{Optane Operation Modes}
The Intel Optane has two operation modes: \textit{Memory Mode} and \textit{App Direct Mode} as Figure \ref{fig:memory_mode_vs_app_direct} shows.
In \textit{Memory Mode} the DRAM works as a cache for the Optane memory. All memory hierarchy  management is made similar to using only DRAM, without any changes to the application. It is worth noting  that applications with a small memory footprint (smaller than the size of the DRAM), will have little impact using this setup. Besides that, the Optane in this mode does not work as a persistent storage device but as a large memory. In the other mode of operation, called \textit{App Direct Mode}, we can use the PMEM as peer memory and perform data persistence. However, unlike \textit{Memory Mode}, the user explicitly defines which part of the application will use the Optane memory.
 
Because \textit{Memory Mode} does not allow the control of which data will be allocated on DRAM or PMEM, we focus on \textit{App Direct Mode}. Our work uses a variant of \textit{App Direct Mode}, supported by the Linux kernel version 5.1, that makes it possible to set up the PMEM as a separate memory NUMA (Non-Uniform Memory Access) node \cite{PMEM_numanode}. In this configuration, PMEM does not use persistence and data is volatile just like it occurs in DRAM. Through memory affinity command, it is possible to decide on what type of memory the object allocations will be made.

\section{Empirical Observations}
\label{sec:motivation}

In a scenario where the entire application cannot be allocated in the DRAM, there is a need  to determine which data will be kept in the DRAM. For this reason, a characterization of the data to be mapped on DRAM is usually performed based on some memory-intensiveness metric. 

\subsection{Baseline: Last Level Cache Misses}

\begin{figure}[h] 
    \centering
    \includegraphics[scale=0.53]{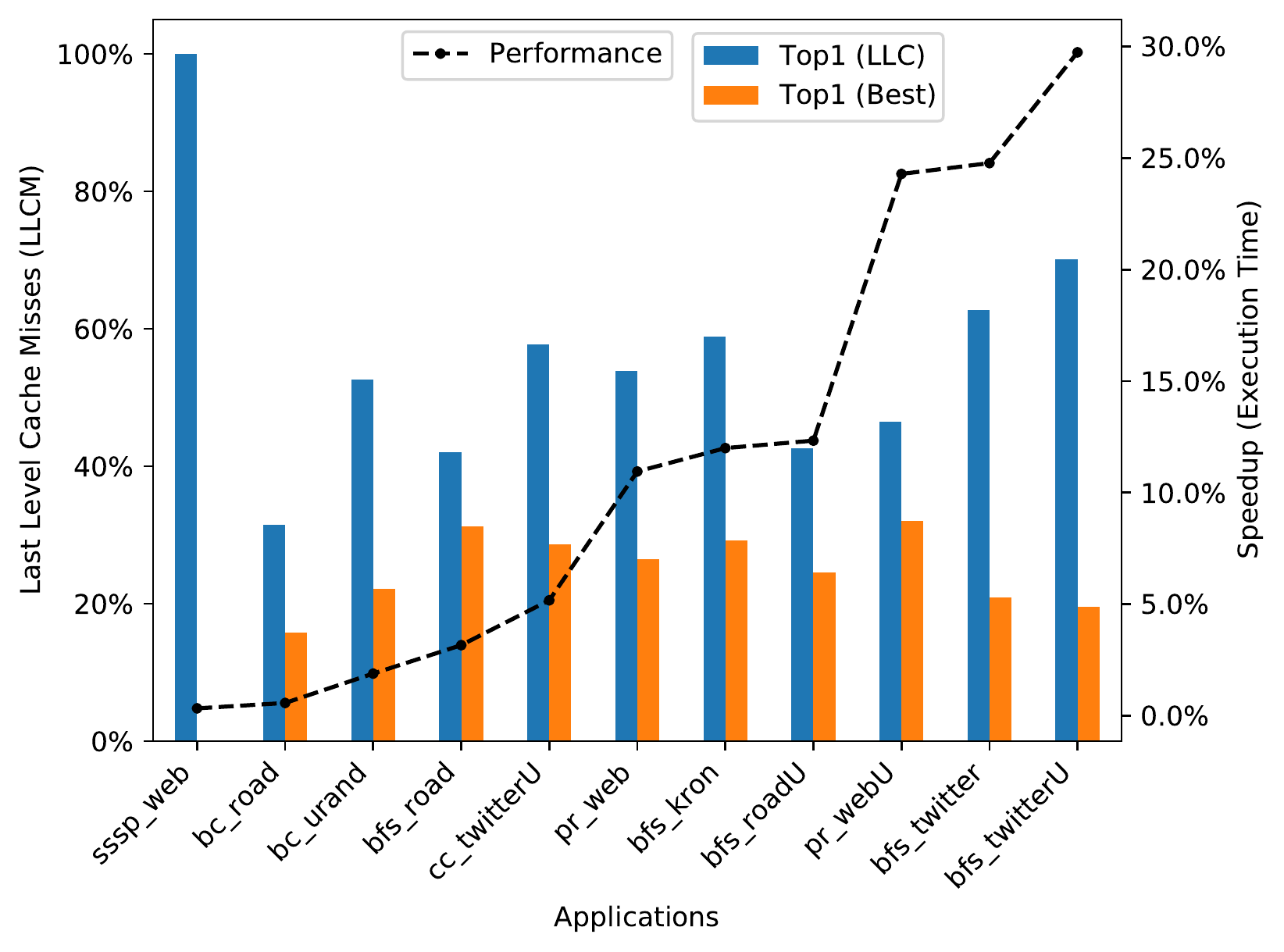}
    \caption{Comparison between the Top 1 object using LLC misses indicator and the best Top 1 object concerning performance (execution time of the application) when allocated in the DRAM.}
    \label{fig:llc_performance}
    \Description{A bar graph showing performance loss when the LLC approach does not find the hottest object.}
\end{figure}

Most works characterize an object as \textit{hotness} using the number of misses in the last level cache. In our work, we consider this approach to be the ``baseline" methodology.
Figure \ref{fig:llc_performance} quantifies how each object speed up the execution time of the application when allocated in the DRAM (more details on the platform used can be found in Section \ref{section:experimental_setup}). The speedup of a particular object is calculated by mapping all objects to PMEM except for the object we want to analyze, which is allocated in the DRAM. The x-axis represents the benchmarks (application + dataset). The left y-axis represents the LLCM for two object types: the one with the highest number of LLCM (\emph{Top 1} - LLC) in the application and the other the object that achieves the highest speedup in the application when migrated to DRAM (\emph{Top 1} - Best). 
The right y-axis shows the speedup of \emph{Top 1} Best over \emph{Top 1} LLC. 
We also order the applications from the smallest to the largest gain.

We can notice in Figure \ref{fig:llc_performance} that, for several applications and datasets, migrating the object with the largest number of LLC misses \emph{Top 1} (LLC) to DRAM did not necessarily bring the best performance. When comparing the two objects \emph{Top 1} (Best) vs \emph{Top 1} (LLC) we see that in all cases there is a difference of at least 10\% in LLCM. In some cases, such as \verb@bfs_twitterU@, this difference is greater than 50\%. The LLC misses indicator does not allow us to infer any relationship with performance. For example, \verb@pr_webU@ the difference is not so great, however, the difference in performance is among the largest in our experiment. This lack of correlation between LLCM and performance indicates that other features must be analyzed to estimate the performance gain when allocating the object in the DRAM.

In the baseline, an object with a high number of LLC misses should be a strong candidate to be allocated in DRAM because its accesses occur outside of the CPU caches, potentially causing a high impact on application performance if PMEM is used. This is how an LLC-misses-based policy works. However, other factors in addition to LLC misses may impact the performance of an object, as will be discussed in Section \ref{subsec:Hotness}.




\subsection{What Really Makes an Object Hot?}
\label{subsec:Hotness}

\begin{figure}[h]
   \centering 
   \includegraphics[scale=0.65]{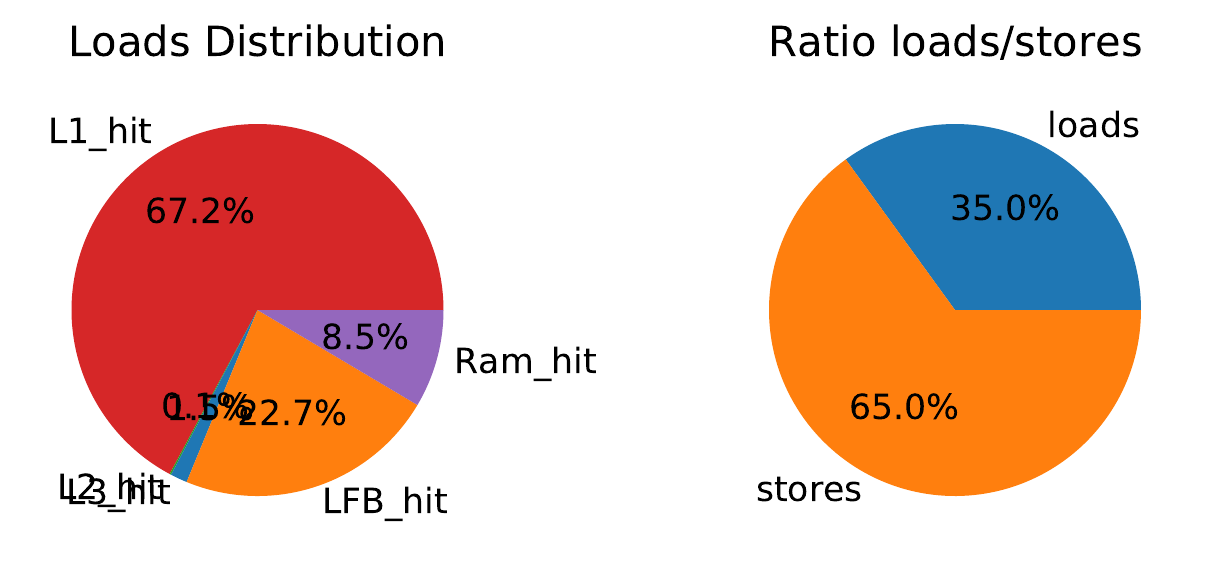}
   \caption{Distribution of load access to different memory levels and load/store ratio to bfs\_twitterU application}
   \label{fig:application_level_overview_bfs_twitterU}
   \Description{A pie chart showing the proportional distribution of the distribution of accesses both in terms of hierarchical memory level and in terms of type of operation.}
\end{figure}

To understand why some application objects with fewer LLC misses end up performing better than others with more LLC misses, we analyze the behavior of the objects in more depth using two applications as example: \verb@bfs_twitterU@ and \verb@pr_webU@. We present an overview of the application \verb@bfs_twitterU@ in Figure \ref{fig:application_level_overview_bfs_twitterU}. The left plot represents the distribution of all loads at different levels of cache and the right plot represents the loads/stores ratio.

By looking at Figure \ref{fig:application_level_overview_bfs_twitterU}, we notice that most load operations occur at the lowest hierarchical level of memory, with 67.2\% in L1 and 22.7\% in Line Fill Buffer (LFB). This demonstrates that the application manages to make good use of its cache. When we analyze the relationship between the number of operations of loads and stores, we notice that this application is store-intensive with almost twice more store than load operations. 
This characteristic is fundamental for a more detailed performance analysis since store operations have a high latency cost when objects are mapped to Intel Optane memory.

\begin{figure}[h]
   \centering 
   \includegraphics[scale=0.65]{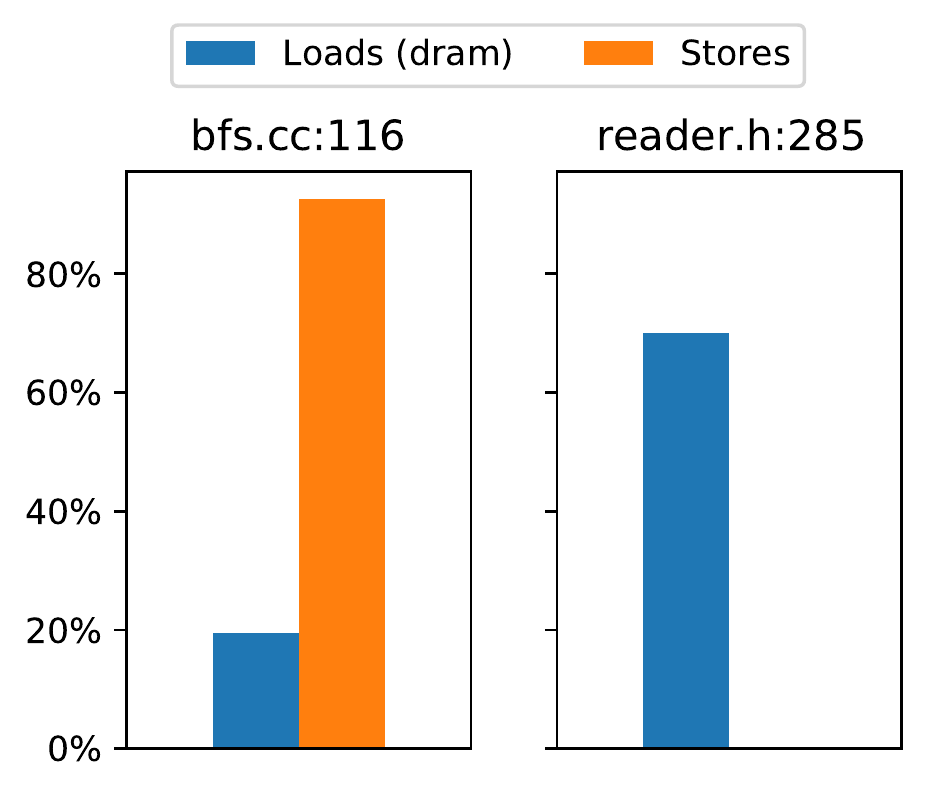}
   \caption{Breakdown percentage of load and stores for bfs.cc:116 and reader.h:285 objects.}
   \label{fig:breakdown_ratio_load_and_store_bfs_twitterU}
\end{figure}

Figure \ref{fig:breakdown_ratio_load_and_store_bfs_twitterU} shows in more details the behavior of two particular objects from  \verb@bfs_twitterU@ application. The object \verb@reader.h:285@ (chosen by LLCM indicator) has around 70\% of all LLC misses and the object \verb@bfs.cc:116@ (\emph{Top 1} -- Best) has around 20\% of all LLC misses. However, even with 3x more load access on DRAM, the application did not perform better when \verb@reader.h:285@ was the object allocated to DRAM. This also suggests that there is a greater number of accesses to DRAM that cannot be accounted for only through the load LLC miss. One of these accesses can occur due to \textit{write-back} operations, either through a load miss or a store miss operation that are preceded by \textit{write-back} operations. In architectures like Cascade Lake, where L3 is exclusive, the DRAM \textit{write-back} policy can occur from a miss that occurs either in L2 or in L3 (LLC). Therefore, considering that the number of writes that occur in the DRAM is more important than the number of reads, since there is a 3x asymmetry concerning bandwidth \cite{joseph2019}, objects with high writes intensity tend to have more impact on the application performance.

As shown by Figure \ref{fig:application_level_overview_bfs_twitterU}, the \verb@bfs_twitterU@ application has a high concentration of store operations. Besides, the object \verb@bfs.cc:116@ has more than 80\% of all stores accesses (see Figure \ref{fig:breakdown_ratio_load_and_store_bfs_twitterU}), while the \verb@reader.h:285@ object is just read-only type. This is a strong indication that store operations are contributing to increased application performance when allocated to DRAM. 

It is important to highlight that not all store operations are converted to a \textit{write-back} operation. The \verb@perf-mem@ profiler reports the number of writes (stores) that occur in the L1 cache level. With the \textit{write-back} cache policy, writes are not immediately mirrored to the main memory. The data in these locations is written back to the main memory only when that data is modified and should evicted from the cache. This makes it difficult to know which of these writes go off-chip. However, we could try to estimate from the access pattern through a feature or use a combination of features to indirectly provide this information.

\begin{figure}[h]
    \centering 
    \includegraphics[scale=0.6]{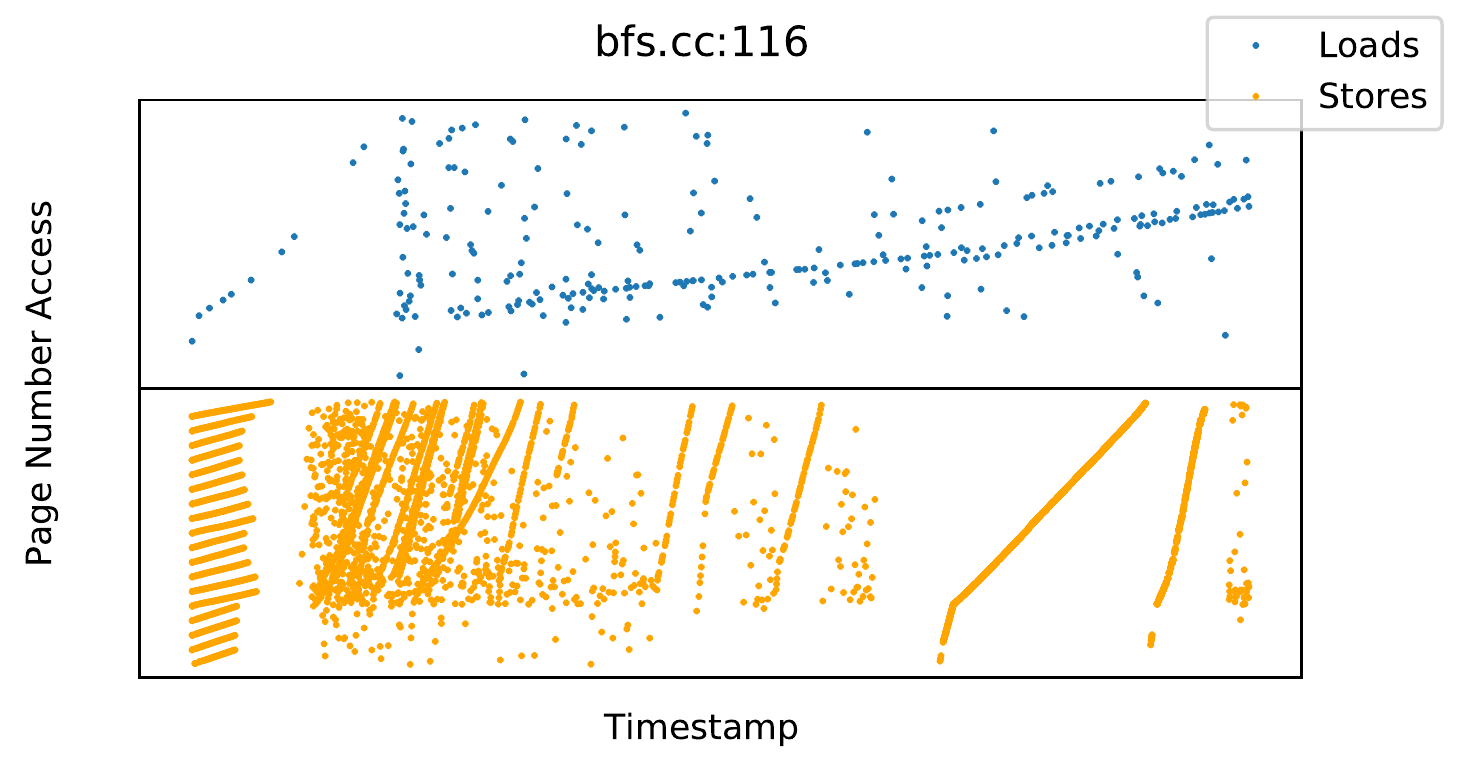}
    \caption{Page access pattern for loads (top half) and stores (bottom half) over time for object bfs.cc:116.}
    \label{fig:object_access_pattern_bfs_twitterU}
\end{figure}

Figure~\ref{fig:object_access_pattern_bfs_twitterU} shows the behavior of L1 accesses for object \verb@bfs.cc:116@. The X-axis represents the timeline and the Y-axis the page number for accesses. The upper part of the plot is the load operations in L1 and the lower part is the store operations in L1.
The operations of loads are more spaced and with a higher level of difficulty in predicting the next access. Store operations appear in greater numbers and show varying levels of difficulty over time  for predicting the next access. 
Our hypothesis is that the observed irregularity (hard-to-predict) in the store operations may be followed by potential load misses and thus can lead to \textit{write-back} operations. This explains the higher performance gain of the object \verb@bfs.cc:116@.

\begin{figure}[h]
   \centering 
   \includegraphics[scale=0.65]{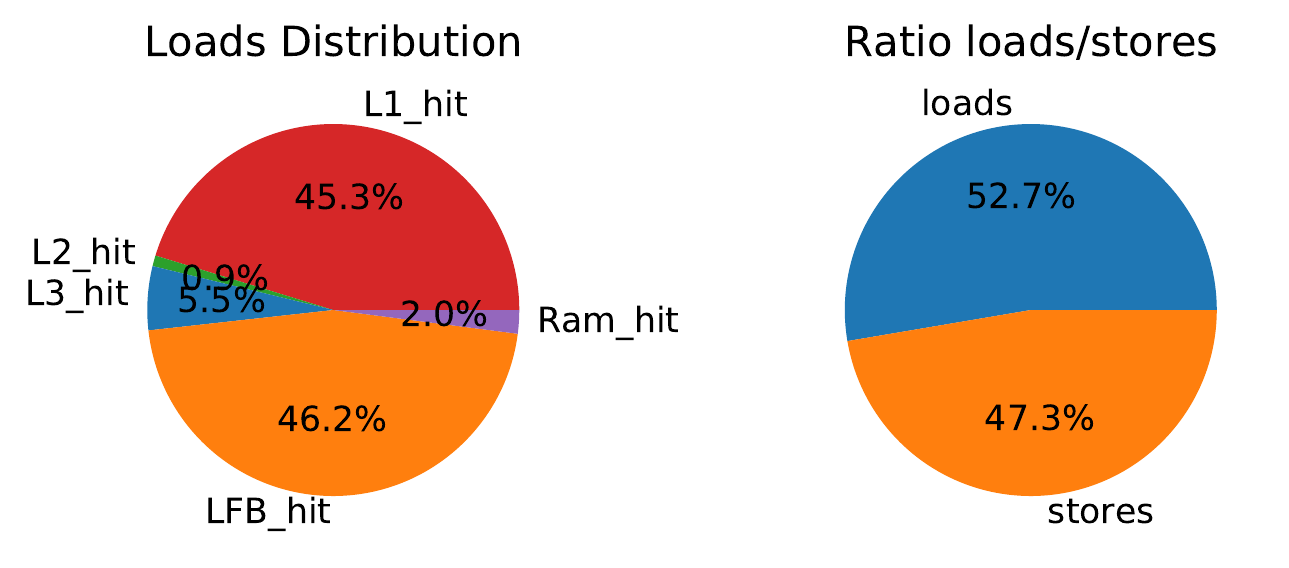}
   \caption{Distribution of load access across different memory levels and load/store ratio for the pr\_webU application}
   \label{fig:application_level_overview_pr_webU}
   \Description{A pie chart showing the proportional distribution of the distribution of accesses both in terms of hierarchical memory level and in terms of type of operation.}
\end{figure}

\begin{figure}[h]
   \centering 
   \includegraphics[scale=0.65]{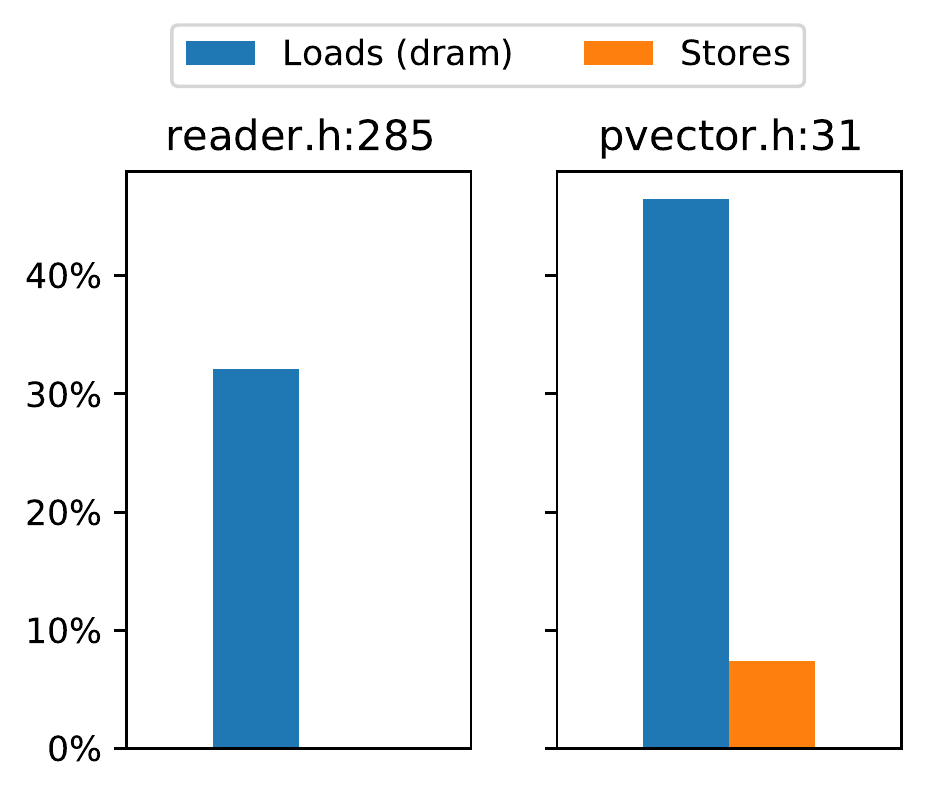}
   \caption{Breakdown percentage of load and stores for reader.h:285 and pvector.h:31 objects.}
   \label{fig:breakdown_ratio_load_and_store_pr_webU}
   
\end{figure}

Figure \ref{fig:application_level_overview_pr_webU} shows the distribution of all loads at different levels of cache and the loads/stores ratio for the \verb@pr_webU@ application. Similar to the analysis of the previous application, most accesses occur in L1 and LFB. However, unlike the previous analysis, the distribution between load and store operations is more balanced, with almost half for each type of access. 
Figure \ref{fig:breakdown_ratio_load_and_store_pr_webU} shows the percentage of DRAM and store (L1) accesses for the \verb@p_vector:h31@ (chosen by the LLCM indicator) and the \verb@reader.h:285@ (\emph{Top 1} best). The latter is read-only and thus there is no influence of store operations that contributes to its performance. The object \verb@pvector.h:31@ (chosen by LLCM) experiences a relatively low intensity of store operations, around 5\% of all the stores from the application.

\begin{figure}[h]
    \centering 
    \includegraphics[scale=0.6]{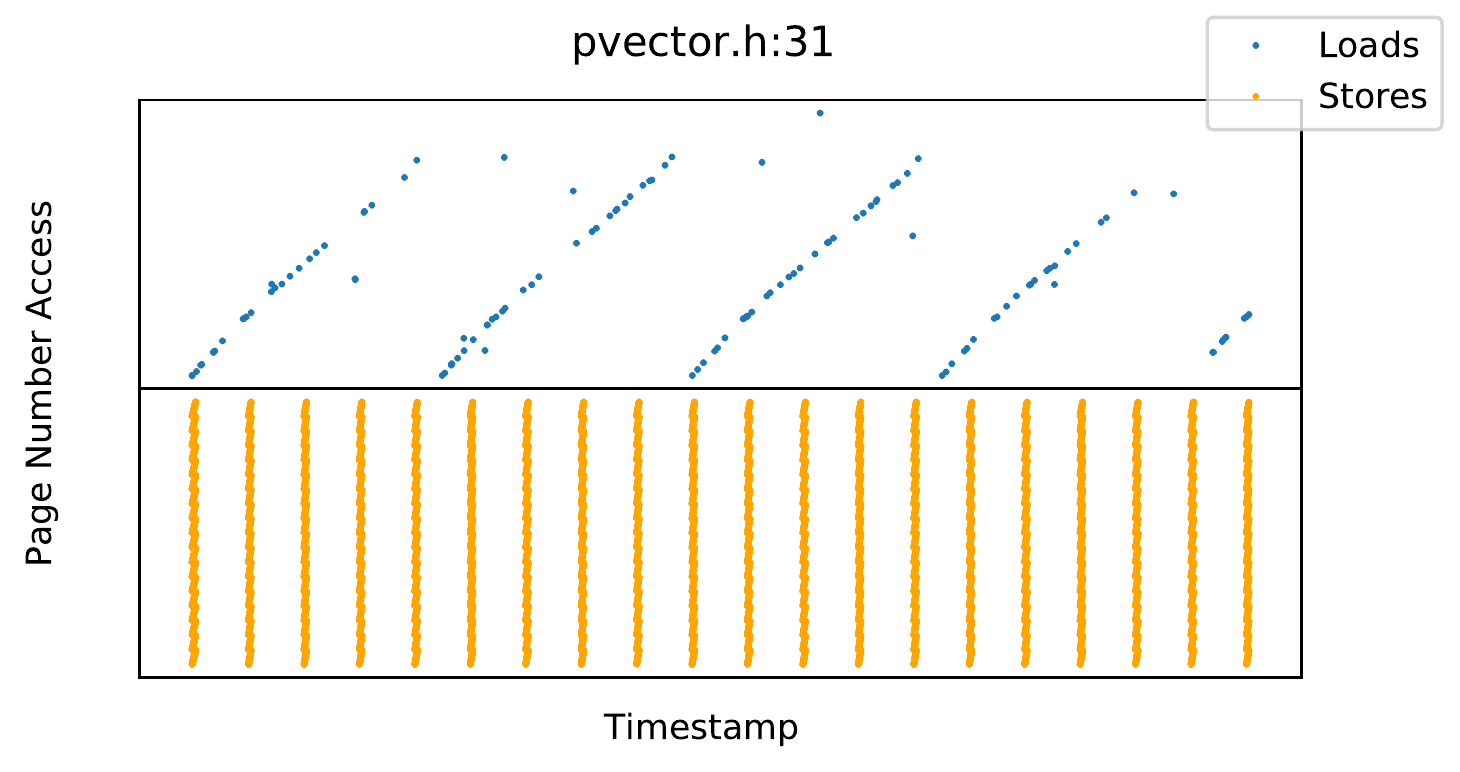}
    \caption{Page access pattern for loads (top half) and stores (bottom half) over time for object pvector.h:31.}
    \label{fig:object_access_pattern_pr_webU}
    \Description{A figure that shows the access pattern of the application's pages.}
\end{figure}

When analyzing the access pattern for the \verb@pvector.h:31@ object, as shown in the Figure \ref{fig:object_access_pattern_pr_webU}, we notice that store operations show a more regular pattern with the next addresses accessed. Our hypothesis is that only a \emph{small portion} of these store operations will be converted into external accesses (\textit{write-back} operations) due to regularity in the addresses accessed. This would justify the low impact of this object when migrated to DRAM.

To help understand why the \verb@reader.h:285@ (\emph{Top 1} best) object performed better, we investigated one more type of operation that performs external access and refers to translation of memory addresses, known as TLBs (Translate Lookaside Buffer). TLBs are responsible for caching the most recently used address translations. They include a subset of the information contained in the \textit{Page Table Entry} (PTE), which is located in the DRAM. Similar to the classification regarding access to data caches, if a TLB lookup finds a match in a TLB, this is known as a TLB \textit{hit}. Otherwise, it is known as TLB \textit{miss}. Our platform (details in Section \ref{section:experimental_setup}) has two levels of TLB, the first being smaller and faster and the second bigger and slightly slower. Both are exclusive to each core. When the address translation is not found at either of these two levels, a search is made at PTE in DRAM. If not found, an event called \textit{Page Fault} occurs to fill the PTE with the requested translation and after such information in the TLB. As a general rule, the further away in the memory hierarchy the address resolution takes place, the longer the time required to access memory data.

\begin{figure}[h]
    \centering 
    \includegraphics[scale=0.7]{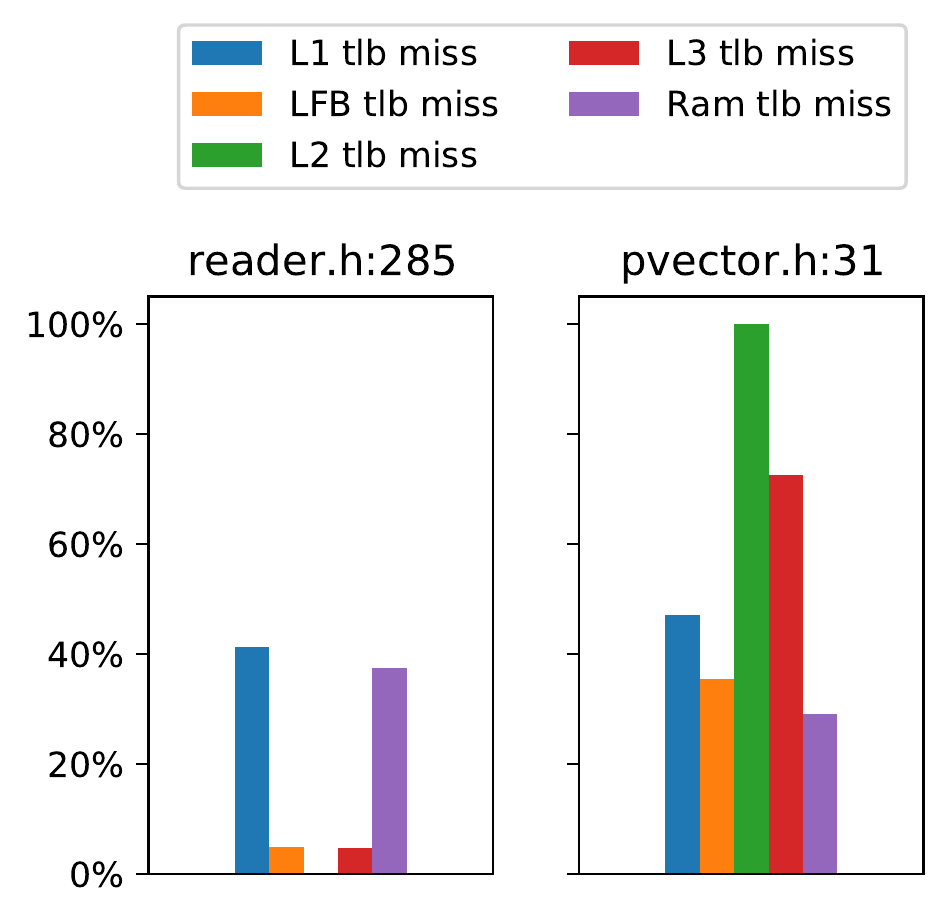}
    \caption{TLB \textit{miss} to different memory levels for reader.h:285 and pvector.h:31 objects.}
    \label{fig:tlb_miss_breakdown_pr_webU}
\end{figure}

Figure \ref{fig:tlb_miss_breakdown_pr_webU} shows the percentage of TLB \textit{miss} for the \verb@pr_webU@ application objects: \verb@reader.h:285@ (Best) and \verb@pvector.h:31@(LLC). At all levels of the memory hierarchy, the \verb@pvector.h:31@ object has a higher value than the \verb@reader.h:285@ object, except for the TLB \textit{miss} that occur in the DRAM. This is precisely the level where exist the highest number of TLB \textit{miss} of the application, with 66.47\% of RAM TLB \textit{miss}, followed by L3 TLB \textit{miss} with 31.71\%. Combined, these 2 TLB levels comprise almost all of the application's TLB miss. Therefore, TLB miss occurring at levels L1 to L2 have little impact on application performance and will not be considered further.

TLB \textit{miss} that occurs in the DRAM tends to be the most costly and likely to be accompanied by a \textit{Page Fault}. 
Therefore, our hypothesis is that because the \verb@reader.h:285@ object has approximately 10\% more TLB miss in the DRAM than the \verb@pvector.h:3@ object, this may have contributed to this performance difference.

The experiments described in this section show that not always a single type of input feature (in this case the ubiquitous LLCM) should decide which object is the hottest concerning overall application performance. In addition, analyzing multiple features together could increase the accuracy of identifying the hottest object. In this work, we propose the use of machine learning models that can best predict the hottest object given all relevant features described in this section.

\section{Design of Rank Hottest Object (RHO)}

\subsection{Profiling Memory Accesses}
\label{sec:memory_profiler}

\subsubsection{Memory Samples}

Since 2012, the \verb@perf mem@ command has been available for sample (not trace) memory access (load, store) through the \verb|perf_events| kernel interface. A memory sample includes the type of event (either a load or a store) and which hierarchical level of memory the event happened. There are several ``load" levels: DRAM, L3, L2, L1 and LFB. Concerning ``store" events, there is only the L1 level. Each sample also includes the data source (address), the timestamp of the sample, the identifier of the thread that issued the memory access, and the latency for the load operations. This memory sampling feature requires (common nowadays) hardware assistance; in our case, we use an Intel Cascade-Lake processor that supports it.

\subsubsection{Tracking Application Objects}
Applications that need high system performance, as in the case of our graph analytics benchmarks, are typically implemented in C/C++ and perform memory allocations using library calls like \emph{malloc} or \emph{new}. These calls are converted to different system calls, depending on the size of the data to be allocated. For small requests the \emph{malloc} is converted to \verb|sbrk| system call. For memory requests larger than the system-defined \verb|MMAP_THRESHOLD| parameter\footnote{In glibc, the default value is set to 128*1024.}, malloc/new uses the \verb@mmap@ system call to find addressable memory space \cite{mmap_size}. As we are working with applications that make use of large chunks of memory (granularity of several pages), we focus on this type of allocation. 

To track memory allocations/deallocations, we use a shared-library  for intercepting Linux syscalls in userspace \cite{Rudoff2017}. For each allocation request (\verb@mmap@ call), our tool stores the base address of the object, its size, and the allocation timestamp.  We also store the context in which the allocation was executed (that is, the call stack information); for this, we use the \textit{backtrace} \cite{backtrace} library. To uniquely identify a \verb|mmap| call, we compute a hash signature using the allocation size, the stack height, and the return addresses (function names plus offsets). The \verb@mmap@ calls that have the same calling context (hash) but are allocated at different times are considered the same object and will not be analyzed in isolation. For example, in case we have two \verb@mmap@ calls with the same hash but are allocated at different times, their associated information (e.g. number of access and lifetime) will be aggregated in a single \verb@mmap@ and added to the dataset.

\subsubsection{Mapping Memory Samples to Objects}
To associate a \verb@mmap@ call with memory access samples, we use memory address and timestamp information from our interface tool with the \verb@perf mem@. For each \verb@mmap@ call, we have a timestamp of its allocation and deallocation; additionally, each \verb@mmap@ includes the range of virtual addresses (extracted during allocation interception). Each memory access sample has the timestamp and the address where the event occurred (derived from \verb@perf-mem@ tool). By combining these pieces of information, we match any sampled memory event to a particular \verb@mmap@ allocation. After all samples have been analyzed and mapped, it is possible to extract statistics at the object level. For example, which object had the highest number of LLC misses, the highest number of stores in L1, and what kind of access pattern. Objects that are allocated and deallocated over time but that have the same call stack are characterized as a single object (all sample data is summed up).

\subsection{Learning to Rank Hottest Object}
\label{subsec:Predictive_modeling}

We introduce a predictive modeling solution responsible for sorting objects based on their influence on application performance (which we call sometimes ``object performance'' for short). For this, we leverage supervised \textit{learning to rank} (LTR) \cite{Li2011}, part of a class of techniques to solve ranking problems. The goal of the LTR is to learn a function from a labeled dataset that allows mapping features to real-valued scores. The function results (scores) are used to order and rank each sample. We do not use traditional classification/regression models because  our LTR model decides based on a list of instances, while traditional models decide on a single instance at a time.

\subsubsection{Ranking Algorithms}
Most LTR algorithms can be grouped into three categories: pointwise \cite{Ordinal_Regression}, pairwise \cite{RankBoost,RankSVM} and listwise \cite{ListNet,AdaRank,LambdaMART}. 
The first two approaches work with one or two inputs at a time and are therefore not so suitable for our context in which we have a list of objects. In the listwise approach, the model receives as input a set of object features and outputs the ranked list. This ranked list is based on scores, which are valid only within each group. 

We notice that LTR algorithms have been replaced by newer algorithms that overcome some noise and bias problems. Among them are the algorithms that are based on trees like the \textit{eXtreme Gradient Boosting} (XGBoost) \cite{XGBoost} and the \textit{Light Gradient Boosting Machine} (LightGBM) \cite{LightGBM} and those that are based on neural networks like TF-Ranking \cite{TF_ranking}. In this paper, we use the tree-based algorithms due to their maturity and better documentation (the second group is still an emerging area of research).

In addition to the advantages already mentioned previously, XGBoost and LightGBM usually present good results because they use a set of weak models that, when combined, form a more accurate model. The models are combined sequentially, where each successor model looks at the errors of the previous model and tries to correct them. Specific algorithms differ in the way their trees are formed. While in LightGBM the tree grows by leaf (leaf-wise), in XGBoost they grow by level (level-wise). These differences impact the model's construction time and accuracy.

\subsubsection{Training/Testing Data}

To generate training and testing data, we used the tools described in Section \ref{sec:memory_profiler}. Each line in the dataset represents an object. From our experiments, objects where its total percentage of operations of loads added to stores does not reach 10\% of the total application loads and stores end up having an irrelevant impact in the context of Graph Applications and can end up hindering the learning of the model; such objects are removed from the analysis. \footnote{This 10\% value was chosen based on our experiments with the characteristics of the analyzed applications. }

We normalize the features associated with each object proportionally to its contribution to the total application features; an object with 20\% of the application's total LLC misses will have a value for this feature of 0.2. The idea is to determine how much each object contributes in different characteristics to the application. In addition to the features of each object, we have also associated the object with the feature of the application. These application features are important and function as a context that applies to all objects in a group. The features we use are  described in Table \ref{tab:features}. We derive new features by multiplying\footnote{We multiply the features given that they are expressed in percentages.} related features at the application level and the object level. For example, if an application has 75\% of store operations (\verb@store_a@=0.75) and an object has 50\% of all these stores  (\verb@stores_o@=0.50), the new feature (\verb@stores@) is the product of these two, storing the value 0.375.

\begin{table}[htbp]
\caption{Application and object features that are derived as input to the model.}

{\begin{tabular}{c|c}
\hline
\multicolumn{1}{c}{\bfseries Level} & \multicolumn{1}{c}{\bfseries Features} \\ \hline
Application & Loads: L1\_{a}, LFB\_{a}, L2\_{a}, L3\_{a}, DRAM\_{a} \\
 & Latency: L1\_{a}, LFB\_{a}, L2\_{a}, L3\_{a}, DRAM\_{a} \\
 & TLB hit: L1\_{a}, LFB\_{a}, L2\_{a}, L3\_{a}, DRAM\_{a} \\
 & TLB miss: L1\_{a}, LFB\_{a}, L2\_{a}, L3\_{a}, DRAM\_{a} \\
 & stores\_{a} \\ \hline
 Object & Loads: L1\_{o}, LFB\_{o}, L2\_{o}, L3\_{o}, DRAM\_{o} \\
 & Latency: L1\_{o}, LFB\_{o}, L2\_{o}, L3\_{o}, DRAM\_{o} \\
 & TLB hit: L1\_{o}, LFB\_{o}, L2\_{o}, L3\_{o}, DRAM\_{o} \\
 & TLB miss: L1\_{o}, LFB\_{o}, L2\_{o}, L3\_{o}, DRAM\_{o} \\
 & stores\_{o} \\
 & mem\_footprint \\
 & group \\ 
 & rank \\ \hline
 
\end{tabular}}
\label{tab:features}
\end{table}

The \textit{group} and \textit{rank} features (last 2 features in Table~\ref{tab:features}) are necessary to define the LTR context. \textit{group} identifies objects that belong to the same application and are important during the cross-validation process. Altogether we have 37 groups. \textit{rank} defines the ranking in terms of impact on performance. In our case, we assume a binary rank, where the top (hottest) object is assigned a value of 1 and the other objects all have a value of 0. The rank is calculated by normalizing the application's execution time when only the object under analysis is allocated in the DRAM over when all objects are allocated in the PMEM.


\section{Evaluation}

We start by evaluating the LLCM approach (baseline), showing the results on different metrics. Then we show how accurate our model is and how much it improves the LLCM approach. We also analyze how each group of features impacts the final accuracy of the models. Finally, we show the impact of carrying out object placement (in DRAM vs PMEM) guided by our predictive model.

Our measurements were carried out through 10 runs and the average was used for each one. The standard deviation min, avg and max are 0.01\%, 0.52\% and 5.42\%, respectively, observed for all executions.

\subsection{Experimental Setup}
\label{section:experimental_setup}

We experiments run on a real machine with hybrid memory: DRAM and Intel Optane. The processor is an Intel(R) Xeon(R) Gold, 18 cores, 192 GB of DRAM (6x32GB DIMMs) and 768 GB of PMEM (6x128GB DIMMs) per socket. To eliminate the influence of non-uniform memory access,  all experiments use only one socket.  In addition, hyperthreading is disabled and the governor is configured as performance (i.e., CPU frequency is set at the maximum allowed). We use six applications described in Section  \ref{section:graphs_application} and 13 real and synthetic datasets with different numbers of vertices, edges, and sizes as shown in Table \ref{tab:datasets}. The synthetic datasets \verb@kron@ and  \verb@urand@ were generated using the original parameters used in GAPBS \cite{gapbs}.

\begin{table}[htbp]
\centering
\caption{Real and synthetic graph datasets used in our evaluation, what include different numbers of vertices, edges and sizes.}
{\begin{tabular}{c|c|c|c}
\hline
\multicolumn{1}{c}{\bfseries Graph Dataset} &
\multicolumn{1}{c}{\bfseries N. Vertices} &
\multicolumn{1}{c}{\bfseries N. Edges} &
\multicolumn{1}{c}{\bfseries Size} \\ \hline
road.sg & 24M & 58M & 806MB\\
road.wsg & 24M & 58M & 1.3GB\\
roadU.sg & 24M & 28M & 403MB\\
twitter.sg & 62M & 1.5B & 12GB\\
twitter.wsg & 62M & 1.5B & 23GB\\
twitterU.sg & 62M & 1.2B & 9.5GB\\
web.sg & 51M & 1.9B & 16GB\\
web.wsg & 51M & 1.9B & 30GB\\
webU.sg & 51M & 1.8B & 11GB\\
kron.sg & 134M & 2.1B & 17GB\\
kron.wsg & 134M & 2.1B & 33GB\\
urand.wsg & 134M & 2.1B & 33GB\\
urand.sg & 134M & 2.1B & 18GB\\ \hline
\end{tabular}}
\label{tab:datasets}
\end{table}

\begin{figure*}[t]
    \includegraphics[scale=0.7]{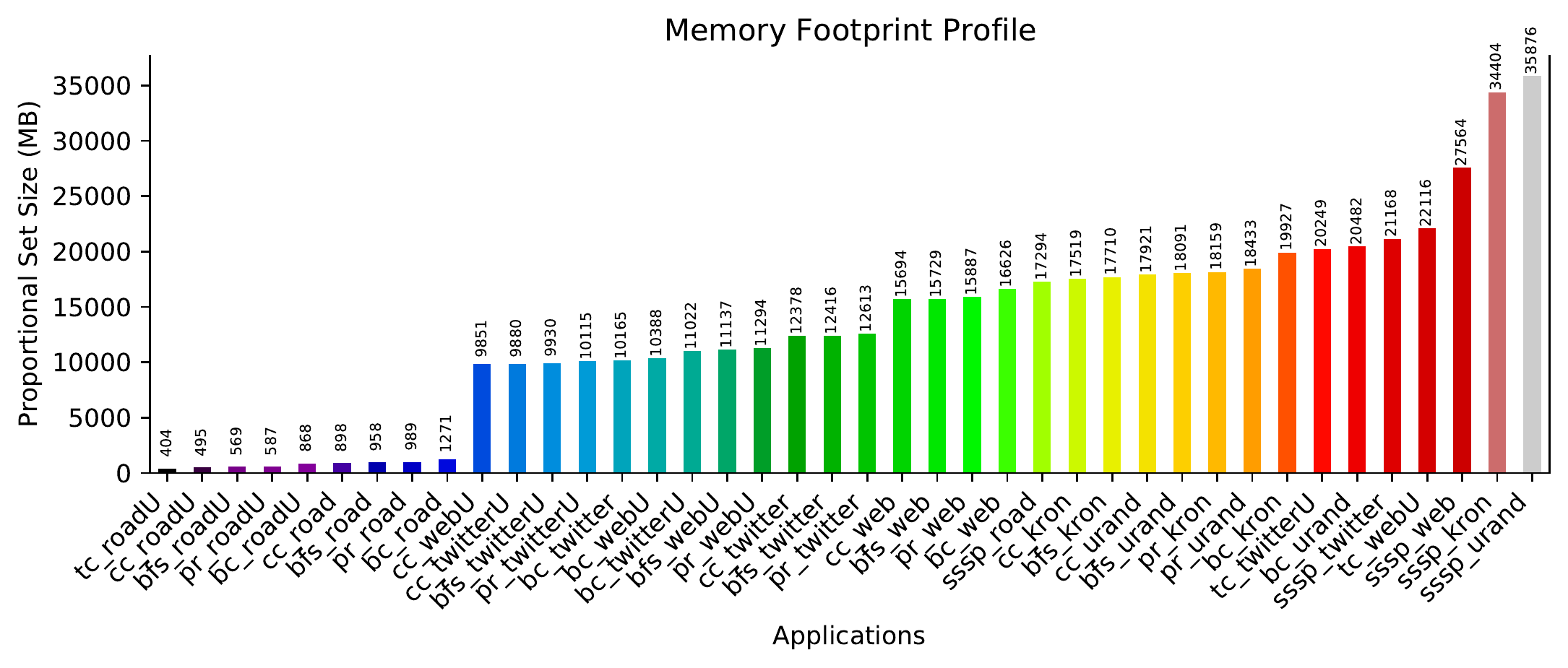}
    \caption{Memory footprint for different benchmarks (application + dataset) using Proportional Set Size.}
    \label{fig:memory_footprint}
    \Description{A set of bars whose height represents the memory footprint of each application.}
\end{figure*}

Figure \ref{fig:memory_footprint} shows the memory footprint for all benchmarks calculated by monitoring the number of pages in memory during execution. More specifically, we monitor the Proportional Set Size (PSS) throughout the application execution; that is, the sum of all private pages plus pages shared by the application process. However, shared pages are only counted in proportion to the number of processes that use them. This metric is a good approximation especially in our case where the vast majority of pages are benchmark exclusive.

\subsection{Model Deployment}

We used the XGBRanker model with the \emph{ndcg} ranking objective function  because it follows the listwise approach. We also use  the following hyperparameter settings: learning rate of 0.03, max depth of 6, and 100 estimators. These values were found from an analysis of hyperparameters (also known as \textit{tuning}). The parameters that brought the best result were chosen. For the LightGBM model, we use the default parameters \cite{lightgbm_api} as they already bring good results in terms of accuracy.

We use k-fold cross-validation with k=37 which is the number of the benchmarks (application + dataset) considered. When the value of \verb@k@ in the k-fold is equal to the total of samples, this approach is called \textit{leave-one-out cross-validation} and is a standard evaluation methodology to evaluate models when sample data volume is relatively small \cite{Loocv} and also to estimate how much the model can generalize for unseen data. 

\subsection{Metrics}

There are several measures (metrics) that are commonly used to judge how well an algorithm is doing on unseen data to different LTR algorithms. Among the main metrics are \textit{Mean Reciprocal Rank} (MRR), \textit{Mean Average Precision} (MAP) and \textit{Normalized Discounted Cumulative Gain} (NDCG) \cite{Mcfee2010}. When multiple levels of relevance are used, NDCG is preferably chosen. However, as in our case, we model the problem as binary judgments, that is, predicting the most relevant object, MRR is the most appropriate metric. The mean reciprocal rank is the average of the reciprocal ranks for a set of requests. If the most important object is placed in position 1, its reciprocal rank is 1, if it is in the second position it will be 1/2, in the third 1/3 and so on. The farther away, the greater its penalty.

We also calculate the \textit{accuracy} that is similar to what happens in classification problems. The calculation is done by dividing the number of times the \emph{Top 1} object of an application was chosen correctly by the total number of analyzed applications., in which the number of times the \emph{Top 1} object was chosen correctly is divided by the total number of applications analyzed. Lastly we calculate the \textit{Accumulated Loss}, which allows us to account for a set of applications the sum of performance losses when the \emph{Top 1} object was chosen wrongly. Basically, every time the \emph{Top 1} (Best) object of an application is not chosen correctly, we calculate the \textit{Performance Loss} for that application in terms of execution time percentage. For example, \emph{Top 1} (Not Best) when mapped to DRAM, in isolation, the application had a runtime of 60 seconds, while \emph{Top 1} (Best) had a runtime of 40 seconds. The performance loss in this case for this application is 50\%.

\subsection{LLC Misses Approach (Baseline)}

The baseline approach that uses LLC misses as a metric to identify the hotness object was implemented using 37 different benchmarks (applications + datasets). In 11 of these applications, the object with the largest number of LLC misses did not perform the best. It is important to note that we only focus on the \emph{Top 1} object. The MRR metric of the LLC approach is 0.84.

\begin{figure}[hbt!]
    \centering
    \includegraphics[scale=0.7]{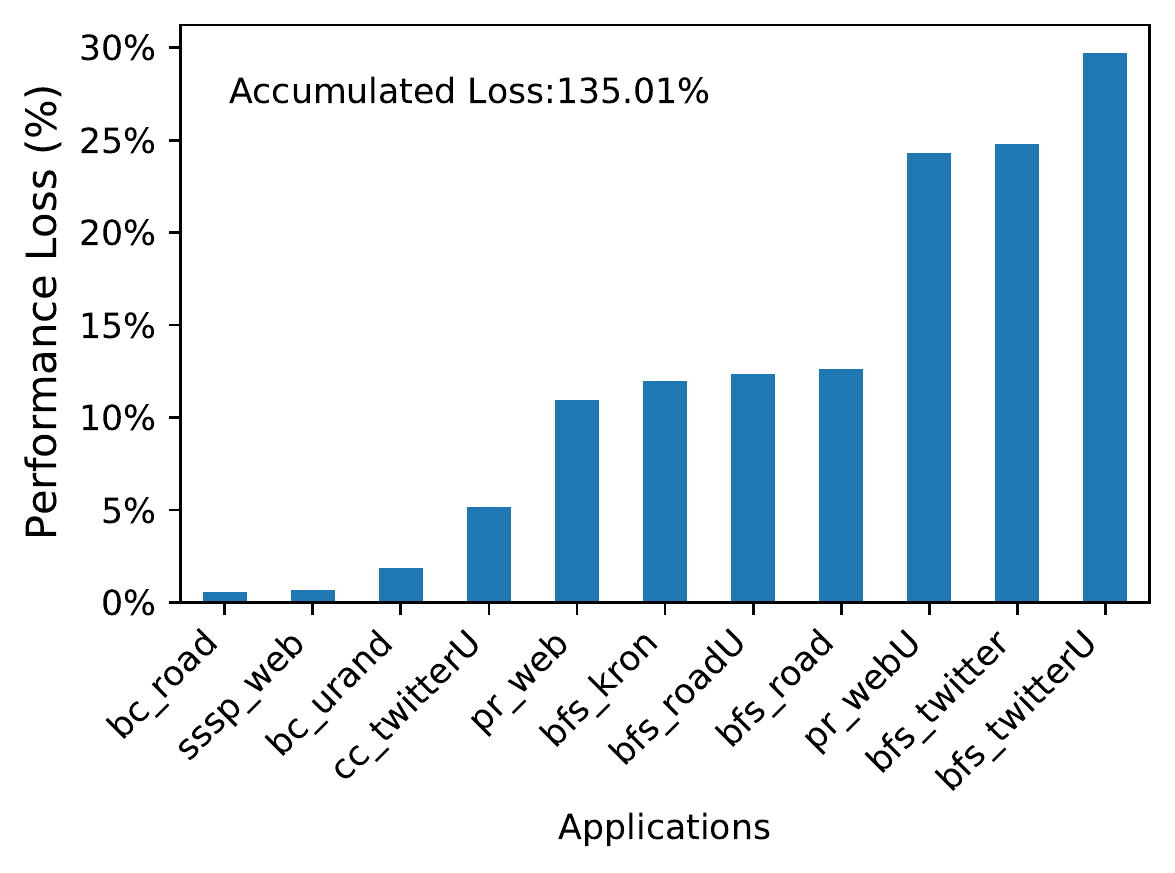}
    \caption{Performance difference for applications in which LLCM approach did not find the best object (Best Top 1).}
    \label{fig:Lost_performance_LLC}
    \Description{The figure shows a set of bars whose height represents the accumulated loss and each bar is represented by an application whose LLC misses approach did not hit the hottest object.}
\end{figure}

Figure \ref{fig:Lost_performance_LLC} shows the applications for which the LLC approach failed to identify the best object to allocate to DRAM.
On the X axis are the applications and on the Y axis is the \textit{Performance Loss}, that is, percentage difference in execution time between the Top-1 using LLCM and the real Top-1 (Best \footnote{The best result comes from analyzing all objects individually at a time in the application. The best is the one that, when allocated in the DRAM, brings the fastest execution time for the application.}). In addition, we calculate the \textit{Accumulated Loss} metric, which is the sum of the losses for each application where the true \emph{Top 1} was not selected correctly, and the LLCM approach lost 135\%. The results also show that in four applications a difference between the LLCM  approach and the best object is small, with differences limited to 5\%. Four other applications have a performance difference of around 12\% and in three other cases the difference are above 25\%.

\subsection{Rank Hottest Object (Our Approach)}

\begin{figure}[hbt!]
    \centering
    \includegraphics[scale=0.7]{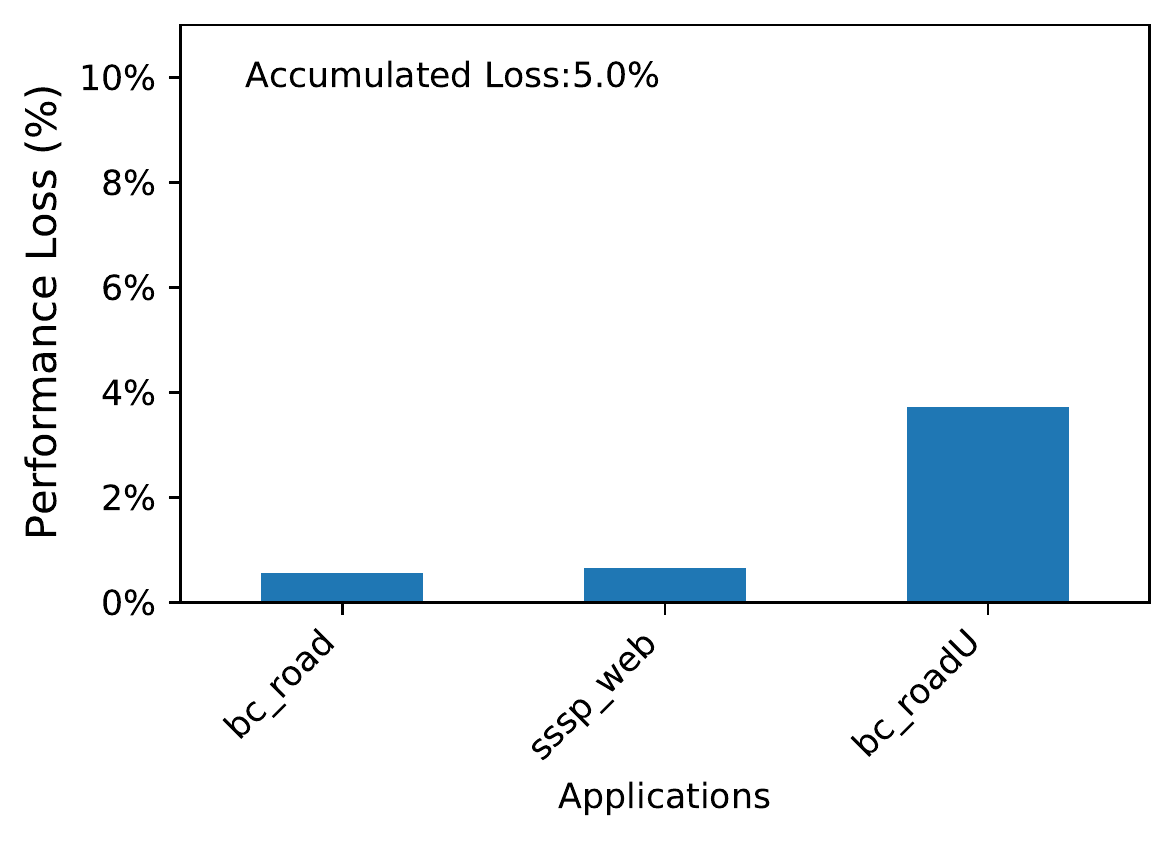}
    \caption{Performance difference for applications in which XGBoost approach did not find the best object (Best Top 1).}
    \label{fig:model_lost_XGB}
\end{figure}

We use two gradient boosting algorithms based on decision trees: XGBoost and LightGBM. The reasons for this choice were explained in Section \ref{subsec:Predictive_modeling}. Our first result was obtained using XGBoost and achieved a \emph{Top 1} accuracy of 0.92. Similar to Figure~\ref{fig:Lost_performance_LLC},  Figure~\ref{fig:model_lost_XGB} shows the (only 3) applications in which our model using XGBoost did not hit the \emph{Top 1} and how much was the difference in terms of performance loss to the best (\emph{Top 1}).

In comparison with the LLCM approach, XGBoost chose more matching objects and when it missed, the performance difference was smaller: the maximum performance loss is 3.4\%. In 2 of the 3 applications where our model did not hit the hottest object, XGBoost placed the hottest object in the second position in terms of performance. With that, XGBoost achieved 0.95 for the MRR metric (note that the maximum value MRR is 1). We also calculate the \textit{Accumulated Loss} and show that XGBoost lost a total of 5\% (recall that LLC misses had accumulated loss of 135\%).

\begin{figure}[hbt!]
    \centering
    \includegraphics[scale=0.7]{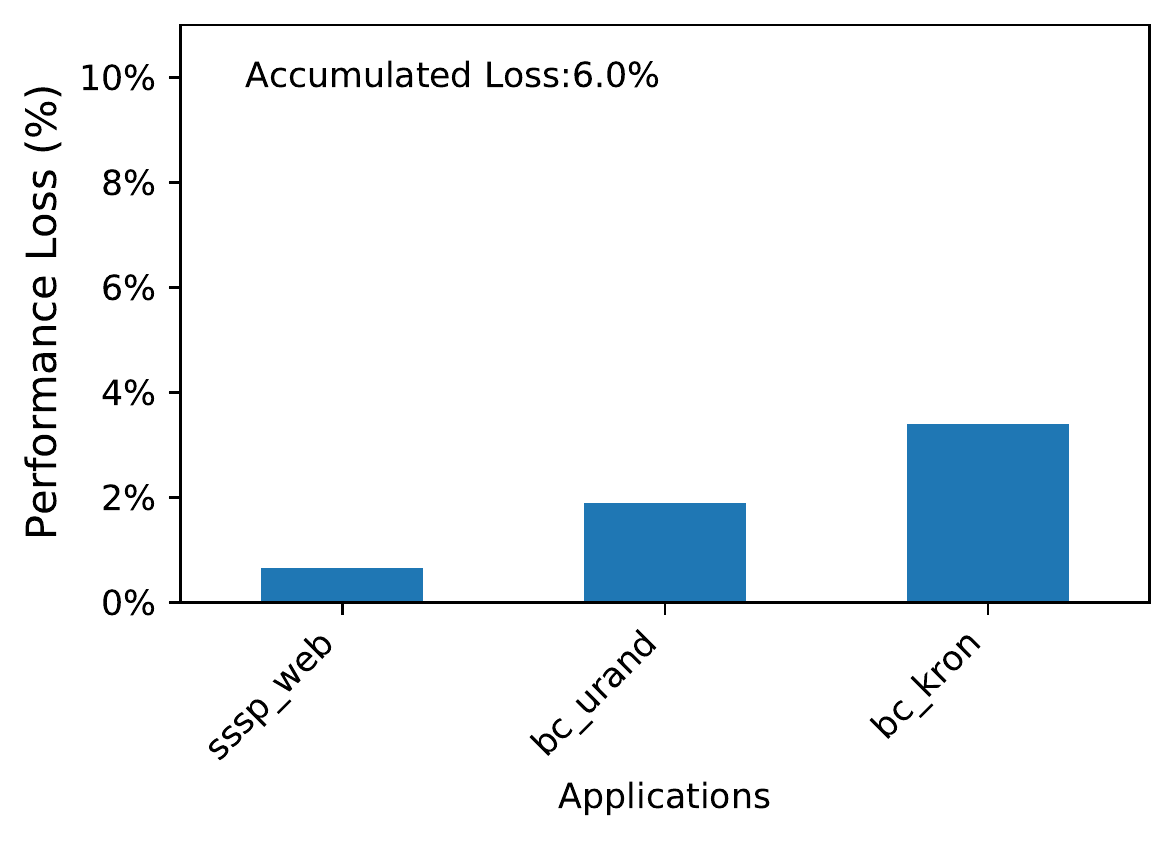}
    \caption{Performance difference for applications in which LightGBM approach did not find the best object (Best Top 1).}
    \label{fig:model_lost_GBM}
\end{figure}

Our second approach using the LightGBM algorithm also achieved an accuracy of 0.92 accuracy for \emph{Top 1} object selection. Figure \ref{fig:model_lost_GBM} shows the (again, only 3) applications in which our model using LightGBM did not hit the \emph{Top 1} and how much the difference was, in terms of performance for the best \emph{Top 1}. LightGBM achieved 0.96 for the MRR metric and reduced \textit{Accumulated Loss} by 6\%.

\begin{figure*}[t]
    \centering 
    \begin{subfigure}{0.75\textwidth}
      \includegraphics[width=\linewidth]{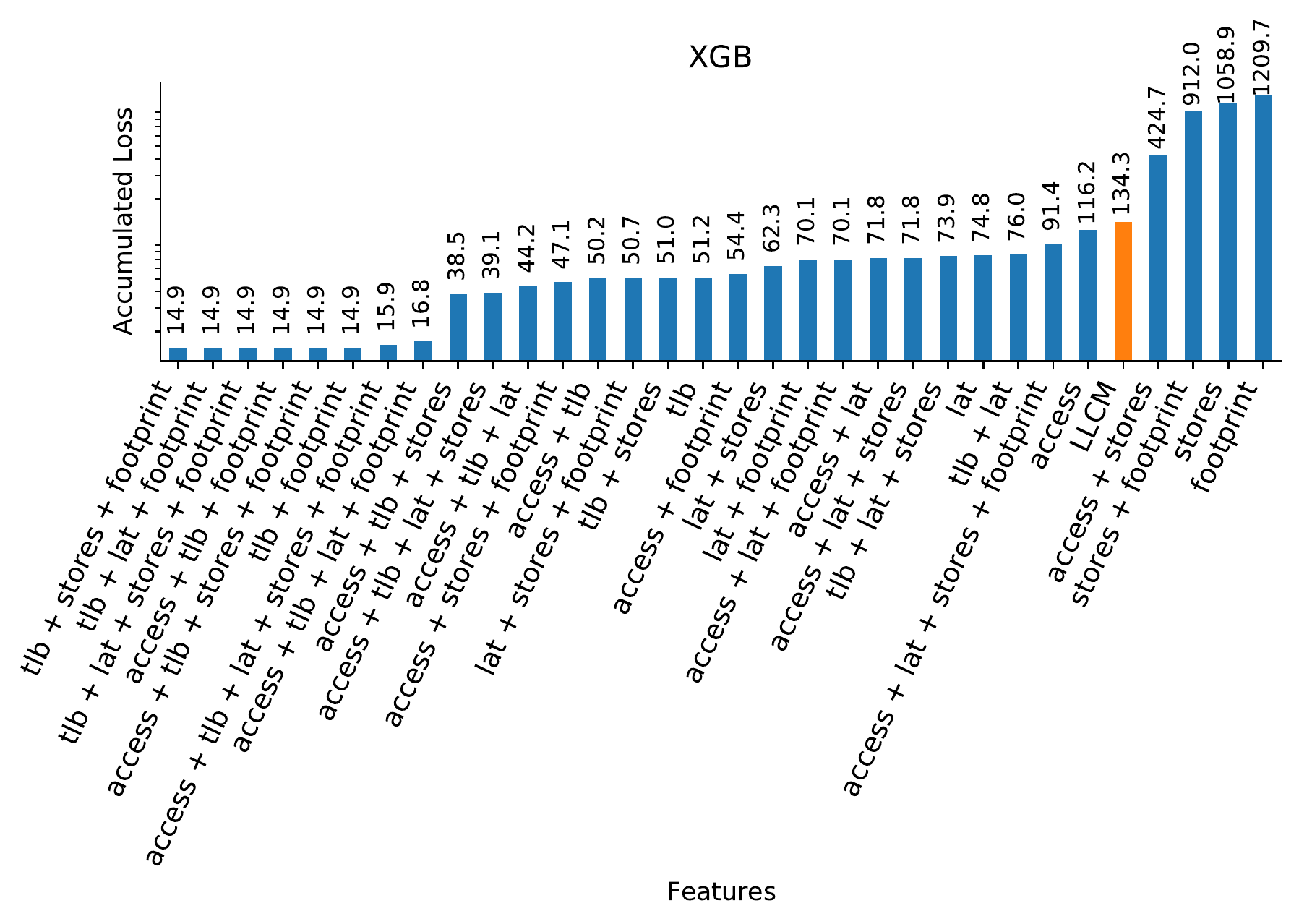}
      \caption{}
      \label{fig:1}
    \end{subfigure}\hfil 
    \begin{subfigure}{0.75\textwidth}
      \includegraphics[width=\linewidth]{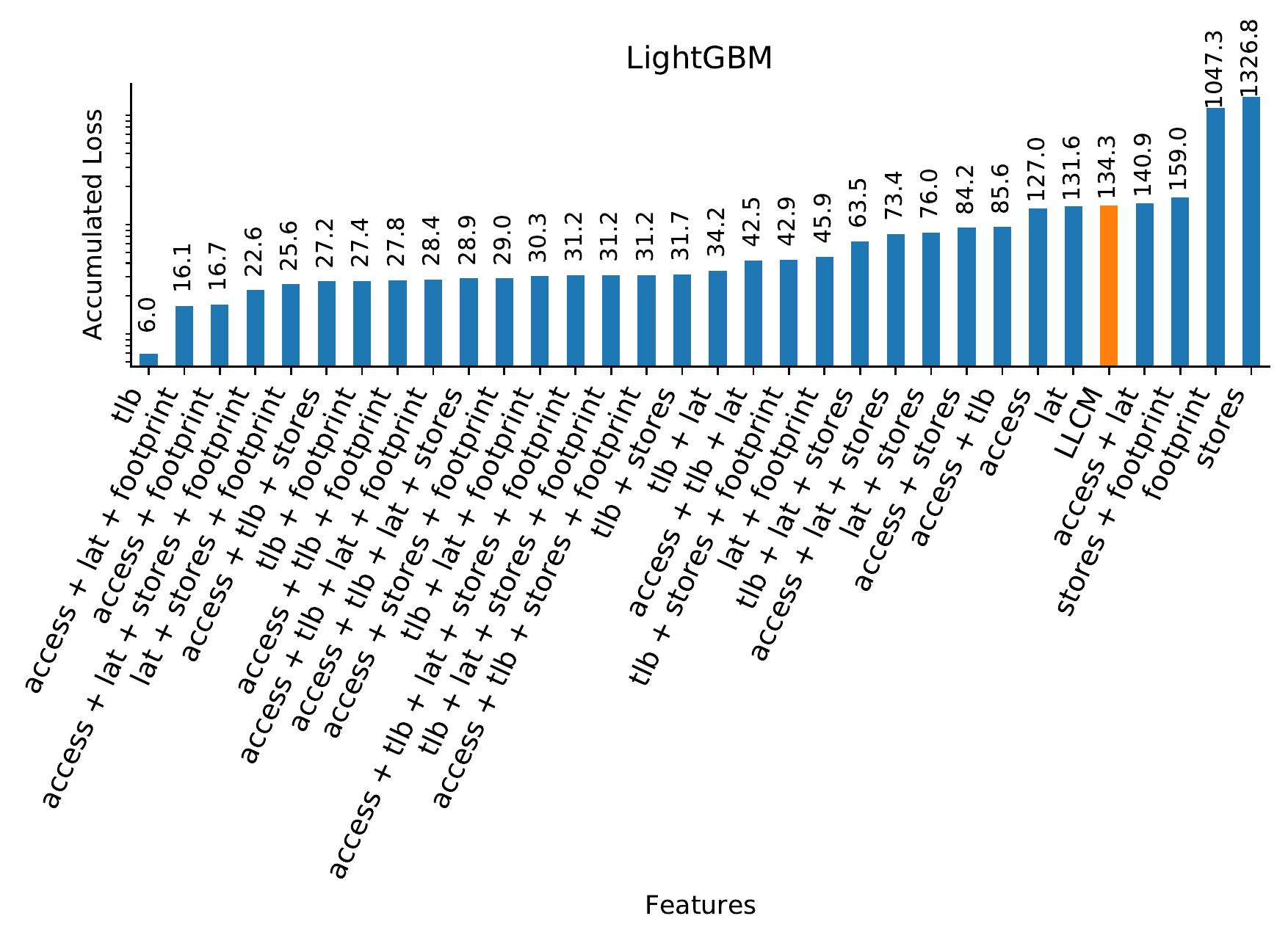}
      \caption{}
      \label{fig:2}
    \end{subfigure}\hfil 
    \caption{Accumulated Loss in function of different features using the XGB and LightGBM models.}
    \label{fig:feature_analysis}
    \Description{The figure shows a set of members whose height represents the accumulated loss and each member is represented by a set of features.}
\end{figure*}

\begin{table}[htbp]
\centering
\caption{Comparative results using different metrics for different approaches.}
{\begin{tabular}{c|c|c|c}
\hline
\multicolumn{1}{c}{\bfseries Approach} &
\multicolumn{1}{c}{\bfseries Accuracy} &
\multicolumn{1}{c}{\bfseries MRR} &
\multicolumn{1}{c}{\bfseries Acc. Loss (avg, std) \%} \\ \hline
LLC & 0.70 & 0.84 & 135 (12.27, 10.18)\\
XGB & 0.92 & 0.95 & 5 (1.60, 1.48)\\
LightGBM & 0.92 & 0.96 & 6 (1.93, 1.14)\\ \hline
\end{tabular}}
\label{tab:final_results}
\end{table}

Table \ref{tab:final_results} shows the summarized results. The accuracy metric only shows whether the approach hit the hottest object or not, among all 37 applications. Sometimes this metric can lead to a wrong conclusion, for example, when the difference between the hottest object and the second hottest object is very small. This is because the difference in terms of performance would be small, but the \textit{accuracy} would count as wrong. The MRR metric considers not only the hottest hit, but when the hottest object is not correctly identified, in which part of the ranking it was placed. Lastly, we show the Accumulated Loss metric (total loss, and average and standard deviation values in parentheses). These values are \textit{calculated based on the number of applications that did not hit the hottest object}, with the total mismatch being 11, 3 and 3 applications for LLCM, XGBoost and LightGBM, respectively. 

Both XGBoost and LightGBM approaches have high accuracy (and high MMR) when finding the hottest object, which was exactly the purpose of our modeling. Our model was only wrong when there was a small difference in performance between the two best hot objects. Because the ranking approach does not aim to match the exact score of each sample, but rather the relative order between the objects of a group, this characteristic of these ranking algorithms favors the discovery of objects with the greatest gain potential related to speeding up execution time. In addition, our modeling (using different types of features across multiple memory hierarchical levels, see Table \ref{tab:features}) may have helped our models to learn the behavior patterns of objects.

\subsection{Importance of Feature Groups}

To understand how each feature in Table \ref{tab:features}  contributes to the final performance of the model, we created different models with different features. The features were grouped into five types: loads (``access"), latency (``lat") and TLB hit and TLB miss (``TLB") at different levels of cache, stores in L1 (``stores"), and memory footprint (``footprint"). 
This grouping was performed by evaluating all possible combinations, one by one, two by two, and so on.

Figure \ref{fig:feature_analysis} shows the results of the feature analysis. The X-axis is represented by combinations of features and the Y-axis represents the \textit{Accumulated Loss}. We show the results in order of \textit{Accumulated Loss}, that is, the combination of features to the left are the better results; we show all single feature types, pairs of feature types, and triple feature types.

The first thing to notice is that each model attaches different importance to different features. For the XGB model, it is possible to see that the information brought by the TLB was fundamental for its learning since for the 10 best results the TLB model was included (See Figure \ref{fig:feature_analysis}-a). On the other hand, for the LightGBM model, although the TLB information brought the best performance for this model, other features were also considered important, such as the footprint and the access (See Figure \ref{fig:feature_analysis}-b). In addition, for the same model, it is possible to learn and hit the hottest object with the same accuracy using different features. For both models, the worst results occur when the model makes use of a single isolated feature. The exception is the use of the TLB feature for the LightGBM model, which ended up bringing the best result. For XGB, the best results always come with more than one type of combined feature.

Sometimes, use all the features available can make learning more difficult if those features do not add significant value to the model. To investigate this issue further, we decided to remove a subset of information from two groups of features: TLB and access. The operations related to address translation (TLB category) can impact the performance of heterogeneous memory, because when a TLB miss occurs is necessary to perform the page walk by traversing the page table in DRAM or PMEM, and/or by taking a page fault. In this case, the type of memory ends up having a significant impact on performance depending on the intensity of TLB miss. To analyze how the information related to the TLB impacts the prediction, we decided to separate the TLB information into two subgroup: TLB hit (L1, LFB, L2, L3, and DRAM) and TLB miss (L1, LFB, L2, L3, and DRAM). The TLB hit subgroup, having DRAM or PMEM as a backend makes no difference since the address resolution has been resolved internally either by the TLB level L1 or TLB level L2. For this reason, we will only work with the second subgroup (TLB miss). In access operations, we also divide into two subgroup: accesses that occur internally (L1 to L3) and external access (DRAM). The first subgroup also has no impact if the application makes use of DRAM or PMEM, while the second does. Thus, we will only work with the second subgroup (External access).

In the case of LightGBM reducing the number of features did not improve the results. Using only the TLB hit and miss information at all levels is the best result obtained by LightGBM. In the XGBoost algorithm the result was different: using only TLB  miss information brought the best result, improving the \textit{Accumulated Loss}, going from 14.9 to 5.0\%.

\begin{figure}[ht]
    \centering 
    \includegraphics[scale=0.75]{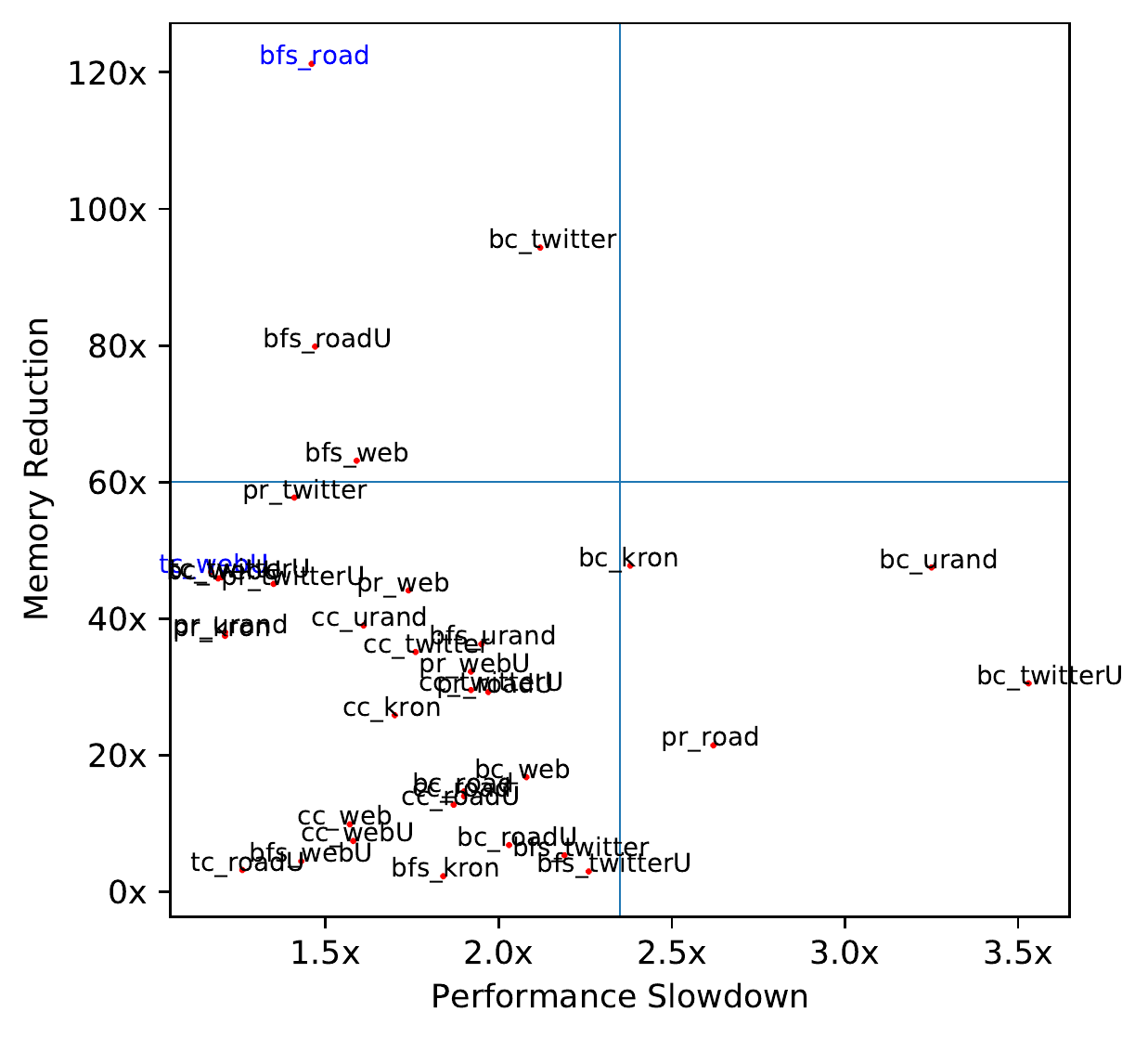}
    \caption{Tradeoff between slowdown and memory footprint reduce for different applications and datasets.}
    \label{fig:dataplacement}
    \Description{The figure shows a Cartesian plane where the x-axis represents performance slowdown and the y-axis represents memory reduction. Inside the plan are the dots representing each application.}
\end{figure}

\begin{figure}[ht]
    \centering 
    \begin{subfigure}{0.29\textwidth}
      \includegraphics[width=\linewidth]{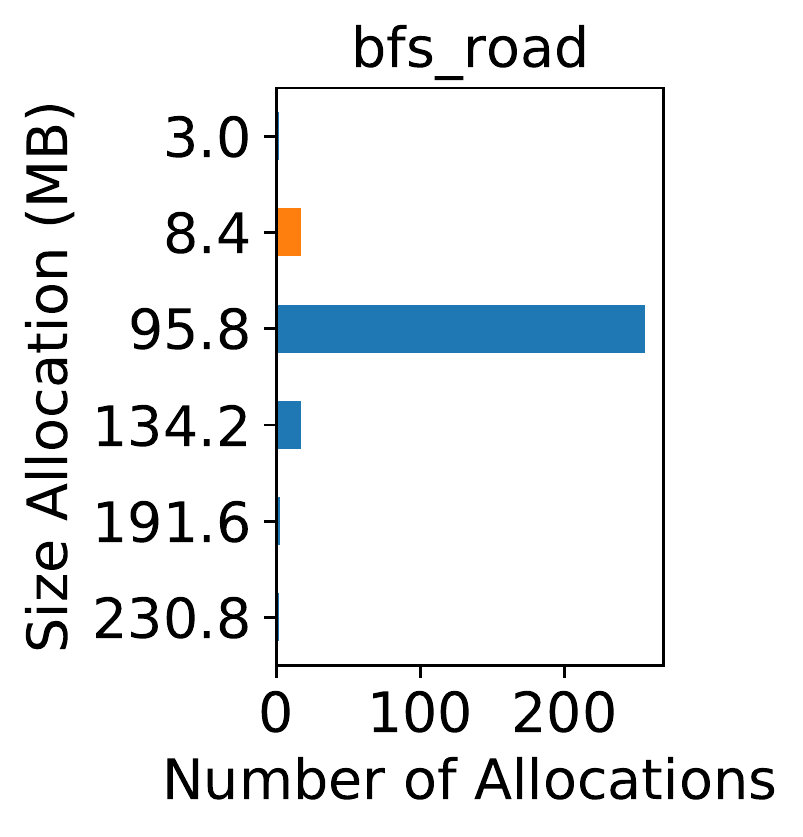}
      \caption{}
      \label{fig:1}
    \end{subfigure}\hfil 
    \begin{subfigure}{0.29\textwidth}
      \includegraphics[width=\linewidth]{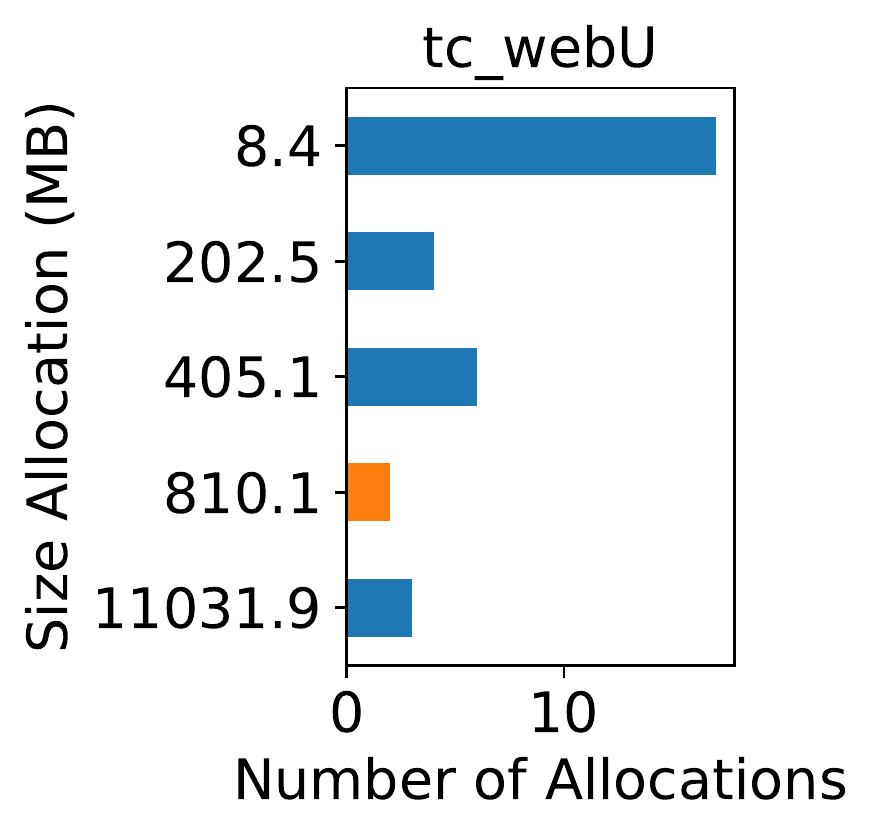}
      \caption{}
      \label{fig:2}
    \end{subfigure}\hfil 
    \caption{Accumulated Loss in function of different features using the XGB and LightGBM models.}
    \label{fig:breakdown_allocations}
    \Description{Two side-by-side bar-shaped figures representing the number of times a given allocation size occurred during application execution.}
\end{figure}

In summary, our experimental results reveal the following insights: 1) modeling the problem with the ranking approach proved to be efficient, since the mistakes made occur in cases where the loss in performance for choosing the wrong hottest object is small. 2) using a single metric as LLCM or Latency does not bring the best result. But also using all the metrics available in our context seems to hinder the prediction. 3) TLB is an important feature in the identification of object performance in the context of graph analytics since this feature is present in the best result of each model.

\subsection{Data Placement of Hottest Objects}

In this section, we will analyze the \textit{tradeoff} between placing only the hottest object (based on our model XGB) in DRAM or placing all objects in DRAM for two metrics simultaneously \textit{Performance Slowdown} and \textit{Memory Reduction}. 

Figure \ref{fig:dataplacement} shows the \textit{tradeoff} between \textit{Performance Slowdown} vs \textit{Memory Reduction}. The X-axis represents the \textit{Performance Slowdown} and is calculated by dividing the execution time when all objects are allocated in the DRAM on the execution time of when only the \emph{Top 1} is allocated in the DRAM. The Y-axis represents \textit{Memory Reduction} and is calculated by dividing the memory consumption when all objects are allocated in the DRAM over when only the \emph{Top 1} is allocated in memory. We also subdivided the plot into four quadrants. The closer to the top and the closer to the left, the better the results are, as it is when the memory consumption is reduced and the slowdown is minimized.

The results show that approximately 31\% of the benchmarks (application + dataset) have up to 1.5x of slowdown when only the \emph{Top 1} object is moved. About 43\% of the benchmarks have degradation between 1.5x to 2x and the other 26\% are the case in which the degradation is greater than 2x. While the slowdown reaches up to 3.5x, allocating a single object also considerably reduces the DRAM consumption up to 121x less compared to allocating all objects in DRAM. This is because in most applications the hottest object is almost always much smaller than the larger object (See Figure \ref{fig:breakdown_allocations}). In summary, we can see that the magnitude of the memory footprint reduction is much greater than the magnitude of the slowdown performance. Thus, it is to be expected that a data placement algorithm may decide to increase  memory consumption to allocate more than a single object in DRAM, in exchange for higher performance gains.

Figure~\ref{fig:breakdown_allocations} shows the different allocation sizes that exist in two applications: \verb@bfs_road@ (best performance slowdown) and \verb@tc_webU@ (best memory footprint reduction). On the Y-axis are the different allocation sizes, while on the X-axis there is the number of times that allocation occurred. The orange bar represents the \emph{Top 1} (best) object. It is important to highlight that two different objects may have the same allocation size and, therefore, this plot does only account for the number of objects of same size. In addition, it does not show which allocations are concurrent and which are sequential. However, it gives us an idea of the size of the \emph{Top 1} object in relation to the other existing objects from application.

In Figure~\ref{fig:breakdown_allocations}-a, the \emph{Top 1} (orange bar) is the second in allocation size and is 27x smaller than the application's largest object. This is probably one of the reasons for this application to have the best result among all 37 benchmarks shown in Figure \ref{fig:dataplacement}  concerning memory savings.  Figure~\ref{fig:breakdown_allocations}-b, the \emph{Top 1} (orange bar) although it is the second largest in terms of allocation size, it is also the object that causes the highest speedup when allocated in DRAM among all \emph{Top 1} of each benchmark. Because of that, this application had the best result in relation to performance slowdown.

\section{Related Work}

Different approaches were previously proposed to solve the problem of data mapping in heterogeneous memories. We categorize prior work techniques along the following lines:

\begin{itemize}
    \item \textbf{Environment}: What was the environment used to evaluate the work? Simulation or Real System?
    \item \textbf{Non-volatile memory (NVM)}: What is the type of NVM? Intel Optane or other?
    \item \textbf{Granularity}: What is the granularity of allocations: Application-level, Object-level, or Page-level?
    \item \textbf{Hottest Data}: Which data is most relevant ? How to calculate? Automatically or manually?
    \item \textbf{Decision}: When does data mapping occur? Off-line systems decide data allocation before the target program runs; on-line systems do it while the program executes.
    \item \textbf{Application/Target}: What kind of application/benchmark was used?
\end{itemize}

Malicevic et al. 2015 \cite{Malicevic2015}, the authors partition performance-sensitive data into DRAM (that is, frequently accessed random data or critical write-only data) and the other data are placed in non-volatile memory technologies (NVM). Dulloor et al. \cite{Dulloor2016} perform a detailed classification for each object based on the size, the percentage of total access, and the percentage corresponding to each access pattern (Sequential, Random, Pointer chasing). Both \cite{Malicevic2015} and \cite{Dulloor2016} show that the frequency of accesses to the application data structures by itself is not sufficient to determine the relative ``hotness" of these data structures. It is also important to know the access pattern. However, both works use an NVM hardware emulator. Therefore, they do not address important implementation aspects such as multi-level caches, pre-fetchers, read/write costs among others.

Similar to our work, Servat et al. 2017 \cite{Servat2017} performs automatic data placement without modifying the source code on systems with hybrid memory. By identifying the objects with the largest number of LLC misses and their respective size, they propose an advisor to report which memory objects are best to place in the fast (DRAM) memory. Once it has been decided which objects will be migrated, the authors use the interposition library  technique to dynamically make allocations. Our approach differs first because we don't classify an object's hotness just by analyzing the LLC. Second, we are  focusing on memory-intensive applications. 

Wang et al. 2018 \cite{Olson2018} propose \textit{MemBrain}, an efficient placement of data in  heterogeneous memory systems. Instead of using LLC misses, they use bandwidth and size of data for each allocation sites (source code file name and line number). This information is obtained using performance counters such as \verb@unc_imc_DRAM_data_reads@ or \verb@mem_load_uops_retired.llc@. Collecting samples every second, they use the average performance counter over the entire execution for each target site. Through an offline process, their approach calculates a ``hotness score'' for each allocation and instruments allocation points in the application code through a custom compilation pass. The idea is to avoid the overhead of context detection during execution. Then, a runtime intercepts these allocations and uses annotations to assign data to different types of memory. However, it is necessary to modify the source code. Furthermore, it differs from our work on the features used, the hot object calculation methodology and the group of applications analyzed.

The \textit{ProfDP}, proposed by Wen et al. 2018 \cite{Wen2018},  is a lightweight profile that quantifies each data object in heterogeneous memories based on latency, bandwidth sensitivity, importance and size. Latency and bandwidth are calculated by running the same application more than once with different numbers of processors. The importance is given by the cost of data access (latency) in CPU cycles. Using these metrics they calculate the Moving Factor (MF), used to describe the benefit of moving a data object to fast memory. Objects with the highest MF have the highest priority to be placed into fast memory. 

The \textit{ATMem}, proposed by Chen et al. 2020 \cite{chen2020}, is an runtime framework for adaptive granularity data placement optimization in graph applications. It works at the level of \textit{data chunk}, that is, each object is divided into data chunks of equal size. They select  critical data chunks as correspond to dense (hot) regions of a data structure that has non-uniform access patterns. It does this using the number of missed reads from the last-level cache LLC as an indicator of priority and normalizes it to the size of a data chunk (Size). During the stage of replacing the chunks, for performance reasons, \textit{ATMem} changes the physical memory of the selected data chunks to high-performance memory without changing the virtual memory address of the data object. One of \textit{ATMem's} problems is that it requires the programmer to indicate, using its own API, when the data migration should occur. Even though the process can be automated by a compiler, it still requires modifying the application. Another problem is that the \textit{ATMem} migrates data during the iterations of graph execution. We argue that this could be best performed once during the object creation of a particular application.

In \cite{Trahay2019}, a tool similar to ours was developed. However, our monitoring tool does not monitor the library call malloc, but the \verb@mmap@ syscall. This considerably reduces the number of intercepts to be managed. In addition, we took it a step further when evaluating the performance of each object when allocated separately in a new type of memory. This makes it possible to assess the impacts that each of them. In \cite{Effler19} the authors have the same purpose as our work, however, they evaluate other features, do not work with memory-intensive applications such as graph analytics and make use of simulation.

What all previous works have in common is that they don't use Intel Optane technologies. Peng et al. \cite{Peng2019} propose two schemes for Optane memory allocation with fine granularity, namely Bandwidth spilling and Write isolation. In the bandwidth spilling scheme, the allocations are divided into blocks, which are placed in CPU sockets in a round-robin fashion. Each block spills from DRAM to NVM. The idea is to distribute the traffic between both types of memory in order to explore the NUMA bandwidth. In the write isolation scheme, the blocks of one data structure are saved into multiple files and then spread over the Optane on two sockets. The objective is to mitigate inter-socket access. The authors do not make it clear how write-intensive data is calculated. Our approach differs from \cite{Peng2019} by performing an offline analysis with more features. In addition, we do not make any changes to the application. \textit{Kleio}\cite{Doudali19} also makes use of machine learning, but using an online scheduler that predicts when a page will be accessed, as well as estimating the hotness of a page using LLCM. Our work differs by using real hardware, working at the object level and focusing on discovering features that impact performance beyond the LLCM feature.

\balance
\section{Conclusion}

Heterogeneous memory has been increasingly adopted given the DRAM limitations and non-volatile memory's density and cost.  The main challenge is determining where to allocate each object, especially important in large-footprint, random access graph-based applications, which are very prevalent in several cloud applications, such as maps and social networks. In this paper, we proposed an offline analysis approach and built models that can predict which object is the hottest using features collected from sampling memory usage in real systems. Our offline approach does not require any application modifications. We show that our model achieves an accuracy of 92\% for the hottest object using real applications and benchmarks widely used in the literature. We also show that it reduces the performance degradation in comparison to the LLCM approach by an average of 12\% and a maximum of 30\% for the applications and datasets analyzed in this work. 
Our model can be extended to rank not only the hottest object but any top \verb@n@ objects. The current model may be applicable, for example, in data placement with heterogeneous memory managing multiple applications where the \emph{Top 1} object of each one is allocated in the DRAM.

\section{Future Work}
We have presented and evaluated an approach to allocating objects in heterogeneous memories using offline analysis. With this approach, an object remains on a single type of memory throughout its lifetime. That is, we currently do not consider object migration. As future work, we will use an online approach, in which the decision of where to allocate each object is decided based on the current state and statistics of the application and system. Also, due to the size of objects being in the order of gigabytes, migrating part of the object can improve memory usage, since most of the time only a part of the pages are used per time interval.

\begin{acks}
We would like to thank all of the anonymous reviewers for their valuable feedback. This study was financed in part by the Coordenação de Aperfeiçoamento de Pessoal de Nível Superior - Brasil (CAPES) - Finance Code 001. The work is also supported by FAPESB (Edital JCB 008/2015) and the Brazilian federal government under CNPq grant (Process No. 430188/2018-8).
\end{acks}

\bibliographystyle{ACM-Reference-Format}
\bibliography{sample-base}



\end{document}